\documentclass[10pt,journal,compsoc]{IEEEtran}

\usepackage[T1]{fontenc}
\usepackage[utf8]{inputenc}
\usepackage{mathtools}
\usepackage[colorlinks,urlcolor=blue,linkcolor=blue,citecolor=blue]{hyperref}
\usepackage[style=ieee,doi=false,url=false,isbn=false,maxnames=99,eprint=false]{biblatex}
\usepackage{listings}
	\lstdefinelanguage{diff}{
    basicstyle=\ttfamily\small,
    morecomment=[f][\color{diffstart}]{@@},
    morecomment=[f][\color{diffincl}]{+},
    morecomment=[f][\color{diffrem}]{-},
  }
\usepackage[svgnames]{xcolor}
  \definecolor{diffstart}{named}{Grey}
  \definecolor{diffincl}{named}{Green}
  \definecolor{diffrem}{named}{OrangeRed}
\newcommand\revision[1]{\textcolor{black}{#1}}
\usepackage{xspace}
\usepackage[inline]{enumitem}
\usepackage{caption}
\usepackage{amssymb}
\usepackage{numprint}
\usepackage{tabu}
\usepackage{multirow}
\usepackage{booktabs} 
\usepackage{dblfloatfix}
\usepackage{siunitx}
\usepackage{amsmath}
\usepackage{mdframed}
\usepackage{makecell}
\usepackage[linesnumbered,ruled]{algorithm2e}

\newcommand{\ie}{\textit{i.e.,}\xspace}
\newcommand{\eg}{\textit{e.g.,}\xspace}
\newcommand{\etc}{\textit{etc.}\xspace}
\newcommand{\etal}{\textit{et al.}\xspace}

\newcommand{\DLTOOL}{\textsf{S4Eq}\xspace}

\bibliography{references.bib}

\begin{document}
%
\title{Self-Supervised Learning to Prove Equivalence Between Straight-Line Programs via Rewrite Rules}
%
%
%
%

\author{Steve~Kommrusch,
        ~Martin~Monperrus,
        and~Louis-No{\"e}l~Pouchet
\IEEEcompsocitemizethanks{\IEEEcompsocthanksitem Steve Kommrusch and Louis-No{\"e}l Pouchet are with Colorado State University, USA.\protect\\
E-mail: \{steveko, pouchet\}@cs.colostate.edu
\IEEEcompsocthanksitem Martin Monperrus is with KTH Royal Institute of Technology, Sweden.\protect\\
E-mail: monperrus@kth.se
}
\thanks{Manuscript submitted 2021-09-21.}}

\IEEEtitleabstractindextext{%
\begin{abstract}
    We target the problem of automatically synthesizing proofs of semantic equivalence between two programs made of sequences of statements.
    We represent programs using abstract syntax trees (AST), where a given set of semantics-preserving rewrite rules can be applied on a specific AST pattern to generate a transformed and semantically equivalent program. In our system, two programs are equivalent if there exists a sequence of application of these rewrite rules that leads to rewriting one program into the other.
    We propose a neural network architecture based on a transformer model to generate proofs of equivalence between program pairs. The system outputs a sequence of rewrites, and the validity of the sequence is simply checked by verifying it can be applied. If no valid sequence is produced by the neural network, the system reports the programs as non-equivalent, ensuring by design no programs may be incorrectly reported as equivalent.
    Our system is fully implemented for one single grammar which can represent straight-line programs with function calls and multiple types.
    To efficiently train the system to generate such sequences, we develop an original incremental training technique, named self-supervised sample selection. We extensively study the effectiveness of this novel training approach on proofs of increasing complexity and length.  Our system, \DLTOOL, achieves 97\% proof success on a curated dataset of 10,000 pairs of equivalent programs.
\end{abstract}

}

\maketitle

\IEEEdisplaynontitleabstractindextext

%
\IEEEpeerreviewmaketitle

\section{Introduction}\label{sec:introduction}

\IEEEPARstart{D}{eep} neural networks have excelled at a variety of classification, reinforcement learning, and sequence generation tasks \cite{goodfellow16}. However, their stochastic nature complicates the use of such networks in formal settings
where one requires a \emph{guarantee} that the result produced is provably correct, such as to assess semantic equivalence between programs.

In this work, we address the problem of proving program equivalence, which is a central problem in computing \cite{kaplan1969regular,godlin2008inference,verdoolaege2009equivalence}. The problem ranges from
undecidable \cite{goldblatt2012well}, to trivial in the case of testing the equivalence of a program with itself. We propose a novel machine learning framework for proving program equivalence, named \DLTOOL, uniquely founded on self-supervision.

\DLTOOL takes as input two programs and generates a sequence of rewrite rules under a well-defined system of semantics-preserving rewrite rules \cite{dershowitz1985computing}.
Our work studies programs represented as a list of statements with straight-line control-flow, using multiple variable types and complex mathematical expressions to compute values.
\DLTOOL outputs a sequence of rewrite rules which is then easily checked for validity. Therefore, only valid sequences are outputted and the system guarantees no false positives by design (no programs are stated equivalent if they are not).
Proving straight-line program equivalence is useful in a variety of contexts: \eg verifying compiler correctness \cite{necula2000translation}, replacing code fragments by more optimized ones \cite{ginsbach2020automatically}, malicious software detection \cite{luo17} or automated student feedback \cite{Clune20}.

We target a challenging instance of the program equivalence problem where the set of rewrite rules we consider may fundamentally change the number and order of operations in the program, such as factorization/distribution or common sub-expression elimination. 
\begin{figure*}
\vspace{-.3cm}
\centering\includegraphics[width=18.25cm]{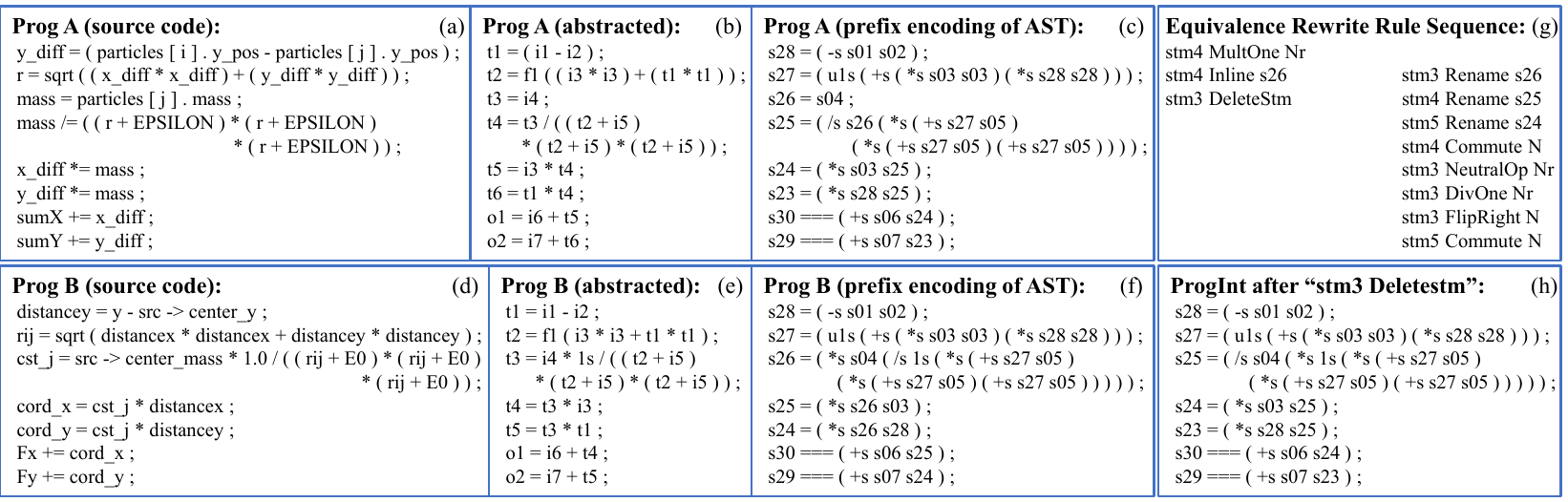}
\vspace{-.5cm}
\caption{Example of various program representations, and application of rewrite rules, in \DLTOOL. Subfigure h shows the intermediate program resulting from applying the first 3 rewrite rules in subfigure g to the program in subfigure c.}
\vspace{-.4cm}
\label{fig:EquivExample}
\end{figure*}
We design a novel self-supervised learning technique, exploiting the ability of our system to automatically synthesize valid new programs and proofs. We initially train a model in a supervised manner with synthetic data which has a broad distribution on the use of rewrite rules. Then we propose a self-supervised technique based on comparing results between broad and narrow proof searches to incrementally train our model. Rewrite rule sequences demonstrating equivalence found by a quick, narrow search are not considered interesting for further training; while sequences found by a broad search indicate samples for which the model's rewrite rule selections could be improved. We name this procedure \emph{self-supervised sample selection}.
We fully implement our learning and inference models in the popular OpenNMT-py framework \cite{opennmt}, based on the transformer model.

To demonstrate the applicability of \DLTOOL on human-written programs, we mined C functions on the popular code hosting platform GitHub \cite{Octoverse}. We collected 13,215 unique programs that we analyze for equivalence by \DLTOOL. This shows that \DLTOOL can prove program equivalence using various compilation and optimization transformations, such as Common Sub-expression Elimination \cite{cocke1970global} that exist on AST samples in the field.
To summarize, we make the following contributions:
\begin{itemize}[noitemsep,topsep=0pt,parsep=0pt,partopsep=0pt,leftmargin=*]
\item We present \DLTOOL, an end-to-end deep learning framework to find equivalence proofs between two  complex program blocks.
\DLTOOL produces a sequence of semantics-preserving rewrite rules to transform one program into the other, via successive rewrites. We consider rewrites which support complex program transformations such as Common Subexpression Elimination and computational strength reduction. \DLTOOL emits a \emph{verified} sequence of rewrites, leading to no false positive by design.
\item We design self-supervised sample selection, an original training technique tailored to our problem domain. This approach further improves the ability of the deep learning system to find more complex proofs.
\item We present extensive experimental results to validate our approach, demonstrating our system can successfully prove equivalence on both synthetic programs and programs derived from GitHub with up to 97\% success. This paves the way for
unsupervised deployment to validate correctness of transformations of program blocks (typical in \eg low-level compiler optimizations \cite{kasampalis2021language}), that are typically unsupported by verification tools \cite{banerjee2014extending}.
\item We provide all our datasets to the community including synthetic generation techniques for the problem of program equivalence via rewrite rules, as well as the programs built from the ASTs we mined from C functions in GitHub \cite{KommruschS4Eq21}.
\end{itemize}

\section{Problem Statement}
\label{sec:background}

We now give the basics of a rewrite-rule based system to prove equivalence between program blocks, the language and rewrites we support, and we finally present theoretical considerations and practical uses of our system.

\subsection{Straight-line Programs}

We target programs made of statements without explicit control-flow instructions (that is, straight-line code \cite{luo17,Ibarra83}). Such programs may be the result of applying full unrolling or flattening on loops (\cite{pai1999code,cong2016source}), or may simply occur in human-written C code as exemplified in Fig.~\ref{fig:EquivExample}(a) and Fig.~\ref{fig:EquivExample}(d).

Our system takes as input programs represented as abstract syntax trees, or ASTs, using specific symbols for input and output values, \eg, \texttt{i1,i2,i3} and \texttt{o1,o2} in Fig.~\ref{fig:EquivExample}(b) and specific typed operator symbols, \eg \texttt{-s} in Fig.~\ref{fig:EquivExample}(c) to represent subtraction of scalars. We require a program to behave as a pure function (no side effect) with a single-entry and a single-exit (SESE), where the set of input and output variable names is known, each distinct name representing a different variable (no aliasing). Under those assumptions, our system  targets proving the equivalence of two programs, represented with their ASTs. The process of abstracting some source code to a safe AST-based representation for equivalence checking is outside the scope of our paper, we assume this process is done beforehand.

For example, Fig.~\ref{fig:EquivExample}(a) and (d) are two actual source code snippets mined from GitHub. The two straight-line programs we generated from them in Fig.~\ref{fig:EquivExample}(b) and (e) do \emph{not} safely capture their exact semantics: live-in variables of different names, \eg \texttt{xdiff} and \texttt{distancex}, are both translated to the same \texttt{i3} symbol in the AST we extracted. In a formal deployment setting, one shall ensure no aliasing between symbols that are non-local to the programs (\eg, \texttt{xdiff} vs. \texttt{distancex}), and compute correspondence between live-in/live-out symbols of each program \cite{zaks2008program}. This is simplified in cases where \eg program $B$ is an optimized version of program $A$, and meant to fully replace $A$ in some larger programs: then by design $A$ and $B$ would always be called with the exact same context. It reduces the entire aliasing and matching problem to simply syntactic name matching.

\begin{table*}
\caption{The 23 rewrite rule categories considered by \DLTOOL. Specializing to different data types supported, 86 rewrite rules are actually considered.\vspace{-.1cm}}
  \label{tab:RulePct}
  \footnotesize
  \setlength\tabcolsep{3pt}
  \centering
  \begin{tabular}{@{}lllll@{}}
    \toprule
 Rewrite Rule and Arguments & Example or Description & & Rewrite Rule and Arguments & Example or Description \\
    \cmidrule{1-2} \cmidrule{4-5}
 SwapPrev & Swap assign statements & &  Inline VarID & Replace VarID with its last assigned expression \\
 UseVar VarID & Replace expression with VarID & & NewTmp NodeID VarID & Assign VarID to expression at NodeID \\
 DeleteStm & Delete assign statement & &  Rename VarID & Change assignment to VarID \\
 AddZero NodeID & $\vec v \rightarrow$($\vec 0 + \vec v$), b$\rightarrow$0+b & &  SubZero NodeID & $\vec v$ $\rightarrow$ ($\vec v$ - $\vec 0$), b$\rightarrow$b-0 \\
 MultOne NodeID & $\vec v \rightarrow$(1$\times\vec v$), b$\rightarrow$1$\times$b & &  DivOne NodeID & a $\rightarrow$ a/1 \\
  Cancel NodeID & ($\vec v-\vec v$)$\rightarrow\vec 0$,(b/b)$\rightarrow$1 & &  NeutralOp NodeID & ($\vec v$ - $\vec o$) $\rightarrow$ $\vec v$, 1$\times$a$\rightarrow$a \\
 DoubleOp NodeID & $- (- \vec v) \rightarrow \vec v$, 1/1/x$\rightarrow$x & & AbsorbOp NodeID & ($\vec v \times$0)$\rightarrow \vec 0$, (b$\times$0)$\rightarrow$0 \\
 Commute NodeID & (a + b) $\rightarrow$ (b + a) & &  DistributeLeft NodeID & (a + b)c $\rightarrow$ ac + bc \\
 DistributeRight NodeID & a(b + c) $\rightarrow$ ab + ac & & FactorLeft NodeID & ab + ac $\rightarrow$ a(b+c) \\
 FactorRight NodeID & ac + bc $\rightarrow$ (a+b)c & & AssociativeLeft NodeID & a(bc) $\rightarrow$ (ab)c  \\
 AssociativeRight NodeID & (ab)c $\rightarrow$ a(bc), (ab)/c$\rightarrow$a(b/c) & & FlipLeft NodeID & -($\vec v$ - $\vec w$) $\rightarrow$ $\vec w-\vec v$ \\
 FlipRight NodeID & a/(b/c) $\rightarrow$ a(c/b) & & & \\
     \bottomrule
  \end{tabular}
\end{table*}

Our input language covers programs made of multiple symbolic expressions using variables of different types, operators, pure function calls, and neutral or absorbing elements (\eg $0,1$). We support both "vector" and "scalar" types, as well as operators and functions that mix these types. We support programs with single or multiple outputs of varying types.

\subsection{Transforming Programs with Rewrite Rules}

We now outline how program transformations are represented in our system. Figure~\ref{fig:EquivExample}(c) and (f) display our actual input program representations, which is strictly equivalent to the programs in Fig.~\ref{fig:EquivExample}(b) and (e). We use a prefix encoding of the AST to feed to our deep learning system presented in later Sec.~\ref{sec:dltool}. The rewrite rules we use operate on this representation.
The parenthesis positioning in this encoding allows for direct recognition of subtrees, and nodes of this tree can be referenced to in the rewrite rules. For example, ProgA in Fig.~\ref{fig:EquivExample}(c) is transformed into ProgB in (f) with the 11 step rewrite rule sequence shown in (g).
The prefix encoding supports operators with 1 or 2 operands and the operator defines the type produced by the operation. For example, \texttt{-s} subtracts 2 scalar values and produces a scalar output.
The input and output languages are fully detailed in our GitHub repository \cite{KommruschS4Eq21}.
Our general rewrite rule syntax in this paper is:\[ \texttt{stm\#  RuleName  [NodeID]  [VarID]} \] where \texttt{stm\#} is the  statement number in ProgA which should have \texttt{RuleName} applied. \texttt{NodeID} optionally identifies the node within the right hand side of the assignment statement, and \texttt{VarID} is the optional name of the variable to use for applying the rule. Precisely, \texttt{NodeID} is a description of the path from the root of the statement to the node of interest, \eg \texttt{Nr}, which is our syntax to model the right child (r) of the root expression node (N). Making the path explicit facilitates learning representations independent of the program size.

For \DLTOOL we have manually specified the rewrite rules summarized in Table~\ref{tab:RulePct}. The 23 different rule groups displayed are specialized for each supported data type, leading to 86 distinct rules. Note we specifically focus on rewrites that go beyond rescheduling operations: we consider rewrites that alter the count and type of operations, such as factorization/distribution, as well as all rules needed to implement symbolic common sub-expression elimination (CSE) on the whole program via a composition of more basic rewrites.

The system of rewrites is how equivalence has been proven in prior work \cite{kommrusch2020equivalence} and we use an identical formalism in this paper. A rewrite rule is made of a tuple <$P,R$> where $P$ is a pattern, or subtree template, dictating when the rule can be  matched on a given subtree of the original AST. $R$ is a rewrite, another subtree template, with which the $P$ subtree is replaced when a successful match occured. To test the validity of applying a rule to a node or subtree in the original AST, we simply need to check the pattern $P$ exists at the expected position in the subtree, if so replacing it by $R$ is valid.

We now illustrate with an example. Taking the fifth assignment in Fig.~\ref{fig:EquivExample}(c), \texttt{s24 = (*s s03 s25)}. We focus on the expression \texttt{(*s s03 s25)}. The rule \texttt{Commute NodeID} from Table~\ref{tab:RulePct} is specialized for all applicable operators and types, in particular the rule $a*b=b*a$ for scalars is defined using <$P,R$> with $P=$\texttt{(*s <subtree-1> <subtree-2>)} and $R=$\texttt{(*s <subtree-2> <subtree-1>)}. \texttt{Stm5 Commute N} is a match: $P=$\texttt{(*s <subtree-1> <subtree-2>)} matches \texttt{(*s s03 s25)} with \texttt{<subtree-1> = s03} and \texttt{<subtree-2> = s25}, the rewritten statement would be \texttt{(*s s25 s03)}. \texttt{Stm5 Commute Nr} is not a match, and therefore an invalid rewrite: the path $Nr$ indicates \texttt{s03}, which is not matched by $P$. Similarly, \texttt{Stm1 Commute N} is not a match, \texttt{-s}, the subtraction of scalars, differs from \texttt{*s}, the multiplication of scalars. \texttt{Stm2 Commute Nrr} is a match: $P$ is found on \texttt{(*s s28 s28)}.

\subsection{Checking the Validity of a Sequence of Rewrites}
Our definition of rewrite rules as <$P,R$> makes trivial the process of checking the validity of a sequence $S$ of rewrites on program $Prog$, and is as follows. For each rewrite $S_i$ in $S$, in increasing order of $i$, do: (a) check $P_{S_i}$ is matched at the node specified in $S_i$. If no, fail: the sequence is invalid. If yes, (b) apply the rewrite $P_{S_i}\rightarrow R_{S_i}$ at the node specified in $Prog$ to generate a new program $Prog'$. (c) Set $Prog = Prog'$. Note this algorithm has essentially a low polynomial worst-case time complexity (pattern size times program size).

It follows our simple procedure to determine equivalence between two programs $P1,P2$ and a candidate sequence $S$ to rewrite $P1$ into $P2$. If the sequence is valid as per the procedure above, and after all rewrites applied on $P1$ the resulting program is \emph{syntactically} identical to $P2$, then the two programs are equivalent. Additional proofs can be found in \cite{kommrusch2020equivalence}.

\subsection{\DLTOOL in a Nutshell}

Our system \DLTOOL operates as follows. For inference (testing whether two programs are equivalent), a pair of programs $P1,P2$ in prefix AST form is input to the system. A maximal length $l$ for the rewrite sequence is set. For no more than $l$ steps, the system (a) proposes a rewrite to be applied; (b) its validity is verified, and if valid, the rewrite is applied to $P1$, generating a new $P1'$ program, which becomes the $P1$ program tested for equivalence with $P2$ at the next step. If at any point $P1$ is syntactically identical to $P2$, the system outputs the full sequence and $P1,P2$ as equivalent. In all other cases, the system outputs $P1,P2$ as non-equivalent.

As extensively developed in Sec.~\ref{sec:datagen} and later, for training, we take as input a grammar for the supported language, and the rewrite rules as defined in Table~\ref{tab:RulePct}. We fully automatically generate new programs and proofs by simply iterating on the grammar productions to generate programs, and iterating on applicable rewrites on a program to generate a pair of equivalent programs and one proof of equivalence for them.

\subsection{Problem Complexity and Relation to Deep Learning}

Given a pair of programs that fit our representation, determining the existence or absence of a sequence of valid rewrites to rewrite one program into another is NP-hard \cite{kommruschphdthesis.21}. Indeed, while some rewrites could be easily handled with a kind of greedy application scheme, such as multOne or Cancel, others immediately create combinatorial explosion, such as Commute. Ibarra and Moran \cite{Ibarra83} discuss the complexity of straight-line program equivalence using a simple grammar: \texttt{z = 1, z = x + y, z = x - y, z = x * y}. They note that, to solve equivalence between such programs, some form of canonical representation could be used (such as sum of products), but as alternating multiplications are distributed over additions this representation can grow exponentially in the size of the programs. While in very specific instances a compact canonical representation may exist \cite{ciesielski2002taylor}, our supported language goes beyond these restricted scenarios and supports arbitrary expressions made of scalars, vectors, \etc for which the problem of equivalence can be proven to be NP-hard. Indeed, the Boolean Satisifiability Problem can be mapped directly into our language such that proving 2 programs equivalent results in solving that classical NP-complete problem \cite{kommruschphdthesis.21}.

Intuitively, we can view the program equivalence solution space $PES$ as a very large graph, where every possible syntactically different program in the language is represented by its own vertex $v$. Then, two vertices $v_i$ and $v_j$ are connected by a labeled edge iff applying one particular rewrite rule on a particular node of $v_i$ is valid, and leads to producing the program $v_j$. The edge is labeled by the rewrite rule applied (as defined above). This graph is a multigraph, as multiple different rewrites may connect the same two programs. It also contains cycles, as a sequence of rewrites can "undo" changes.
Therefore, any two programs connected by a path in this graph are semantically equivalent: the rewrite sequence is the set of labels along the edges forming the path. Additionally, given a case where multiple valid rewrite rules are available at vertices on the path, the number of programs reachable by following edges in this graph grows exponentially with path length and the problem of finding a path in this graph between two programs is NP-hard.
Building the rewrite rule sequence $S$ for $ProgB \equiv S(ProgA)$ amounts to exposing one path (out of possibly many) from $ProgA$ to $ProgB$ in this graph when it exists, the path forming the proof of equivalence. In this work we build a deep learning system to learn a stochastic approximation of an iterative algorithm to construct such feasible path when possible. In a nutshell, we view program equivalence as a pathfinding problem in $PES$, and train a neural model to find efficiently paths in $PES$, only by sampling from $PES$ at training time. Our approach avoids entirely the need to craft smart exploration heuristics for such a large equivalence graph to make this path-finding problem practical (akin to building tactics in theorem provers): instead, the traversal heuristic is learned automatically, without any user input, by deep learning. 

As detailed in later Sec.~\ref{sec:dltool}, one fundamental contribution of our work is to design a self-supervised incremental training approach to address this pathfinding problem. It is motivated by the fact that the \emph{in-practice} complexity of proving equivalence may vastly differ between two problem instances. Our system identifies instances that were not solved well, builds a family of new problem instances derived from it, and incrementally trains on those to improve its abilities on complex cases. Implicitly, instances with high in-practice complexity that are not solved well will be isolated and used for refining the system's capabilities.

\subsection{Applications of Equivalence Proof Search}

Our system proves equivalence between symbolic straight-line programs. We have exemplified on a two-type language with a standard operator mix, up to token renaming, which covers instances of SIMDized functions, register-level code, \etc
As such, \DLTOOL can be used to address the verification and certification of certain compiler optimizations \revision{such as common subexpression elimination, strength reduction, register reuse, and statement reordering} \cite{Muchnick97, 10.1145/358438.349314}. While numerous approaches based on abstract interpretation address equivalence under \emph{reordering} of operations, \eg \cite{verdoolaege2009equivalence,karfaaffinetcad2013,bao2016polycheck}, they typically cannot handle transformations at the statement level. \DLTOOL can take blocks in two programs and test them for equivalence, (re)discovering which statement corresponds to which in the transformed program, helping to address the well-known statement matching problem. Because \DLTOOL can reorder statements, inline variables, and create new temporaries during a proof, it can prove equivalent a full-unrolling of a code region, as is typical in high-performance code generation \cite{kong2013polyhedral}, a fundamental problem in compiler certification. \revision{Proving compiler optimizations correct can be addressed using transfer function graphs \cite{Dahiya2017}, however it requires time-consuming SMT solving to build the equivalence proof.
Luo \etal \cite{luo17} also shows that the ability to detect straight-line equivalence can be combined with other techniques to create obfuscation-resilient similarity detection which can be applied to plaigarism as well as to malicious software detection.}

\revision{The equivalence problem we target is also related to more generally proving mathematical theorems, with recent successes in applying deep learning to find proofs automatically \cite{alhussein19,bansal19,paliwal19}. In this work we target a problem-specific instance of tree rewriting, which is applied to proving the equivalence of ASTs from straight-line code programs, and introduce self-supervised sample selection to address this problem.}

Finally, another use case is the equivalence between linear algebra formulas, as could typically be found in student exercises. \revision{In fact, when including matrices in our supported input language, \DLTOOL is able solve all the matrix expression equivalence programs from 2 relevant Khan academy modules designed to test student’s knowledge of matrix algebra \cite{kommrusch2020equivalence,khan20}.}

\section{\DLTOOL: Deep Learning to Find Rewrite Rule Sequences}
\label{sec:dltool}

We propose to use a deep learning model to find rewrite rule sequences which transform one program into a semantically equivalent target program. The idea is to first learn from correct rewrite sequences, and then learn to solve previously unseen program equivalence problems.

\subsection{Overview of \DLTOOL}

Prior work has shown that source code has patterns that are similar to human language \cite{hindle2012naturalness}, and thus techniques used in natural language processing can work on source code as well, including deep learning \cite{chen2019sequencer}. Deep learning has the ability to learn syntactic structure and expected outputs related to programs.
For \DLTOOL we aim to create a deep learning model which, given 2 programs, will predict a sequence of rewrite rules which can formally prove equivalence between the 2 programs.

For \DLTOOL we use the state of the art sequence-to-sequence deep learning model known as the transformer model \cite{sutskever2014sequence}. Because sequence-to-sequence models are stochastic, they can be used to produce multiple answers for the same query; this is called \emph{beam search}. By using beam search we can order rewrite rules proposed by the model. By composing beam search recursively, we can construct proofs which use heuristics learned by the model to guide proof search.

In \DLTOOL, we devise an original training process specifically for the task of proving equivalence. The \DLTOOL training process combines supervised and self-supervised learning and does not require human labeling.

\begin{figure}
\vspace{-.4cm}
\centering\includegraphics[width=8.8cm]{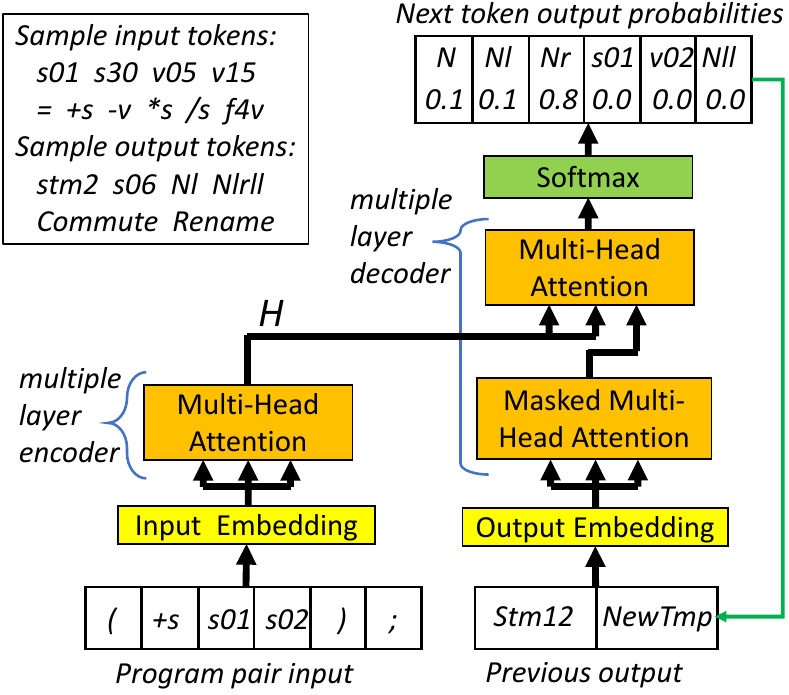}
\caption{Transformer model used to predict rewrite rules given 2 input programs.}
\vspace{-.4cm}
\label{fig:Transformer}
\end{figure}

\subsection{Transformer Model}
\label{sec:transformer}

The transformer sequence-to-sequence model is meant to learn mappings between two sequences, typically of words \cite{sutskever2014sequence}. It is widely used in automated translation \cite{wu2016google} and text summarization \cite{nallapati2016abstractive}. Sequence-to-sequence models consist of two parts, an encoder and a decoder. The encoder maps the input sequence $X = (x_{0},x_{1},...,x_{n})$ to an intermediate continuous representation $H = (h_{0},h_{1},...,h_{n})$, also known as an embedding. Then, given $H$, the decoder generates the output sequence $Z = (z_{0}, z_{1},...,z_{m})$. The size of the input and output sequences, $n$ and $m$, can be different; and the languages for $X$ and $Z$ can also use different vocabularies. A sequence-to-sequence model is optimized on a training dataset to maximize the conditional probability of $p(Z \mid X)$, which is equivalent to:

\begin{align*}
\begin{split}
p(Z \mid X) &= p(z_{0},z_{1},...,z_{m} \mid x_{0},x_{1},...,x_{n})\\
            &= \prod_{i=0}^{m} p(z_{i} \mid H, z_{0},z_{1},...,z_{i-1})
\end{split}
\end{align*}

For \DLTOOL, we introduce the input and output languages for $X$ and $Z$ in Section~\ref{sec:background}.
Subfigures~\ref{fig:EquivExample}(c) and~\ref{fig:EquivExample}(f) are examples of such inputs.
The output language for $Z$ is illustrated in subfigure~\ref{fig:EquivExample}(g).
For input to the transformer, we add a special token \texttt{Y} to separate the 2 programs, with ProgA being input before ProgB.

In \DLTOOL , the transformer model we use is shown in Figure~\ref{fig:Transformer}. The yellow boxes represent the model's learned interpretation for the tokens in the input and output. Tokens such as '*s' and '=' in the input language or 'stm3' and 'Commute' in the output language have learned embeddings used by the transformer model. The model accomplishes context-dependent interpretation with multiple attention layers which learn a complex representation for a program node by learning which other nodes should be ``attended to'' while creating the higher level representation.

\begin{figure*}[h!tb]
\vspace{-.1cm}
\centering\includegraphics[width=18.3cm]{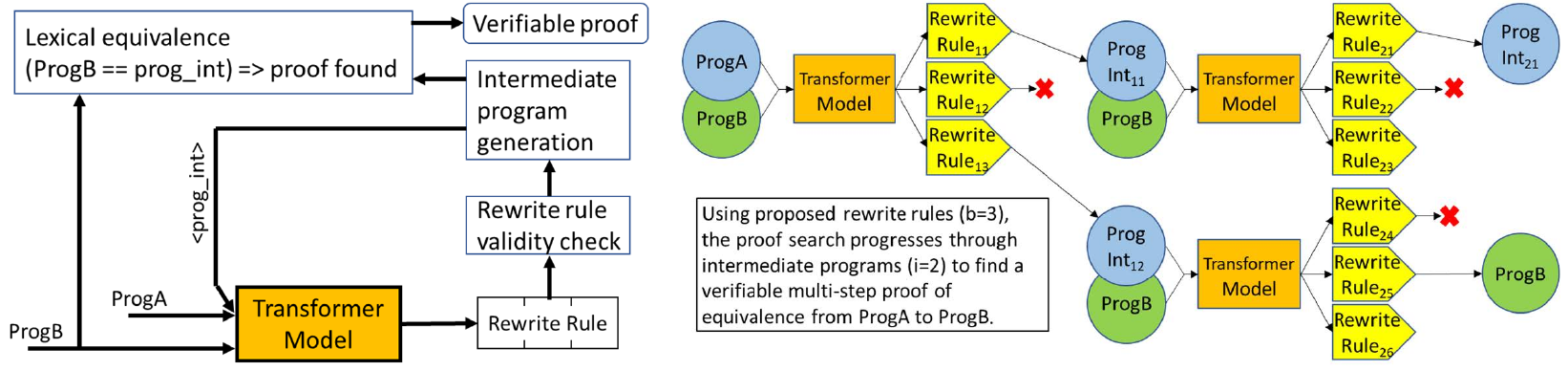}
\caption{Proof search: Transformer model produces multiple rewrite rules at each step. Invalid rules are discarded, lexical equivalence is checked, and up to $i$ intermediate programs are used to produce new rules for the next search step.}
\vspace{-.6cm}
\label{fig:ProofSearch}
\end{figure*}

In the transformer model, the encoder and decoder use layers which provide an attention mechanism which relates different positions of a single sequence in order to compute a representation of the entire sequence. For example, the attention layers provide a mechanism for information related to the assignment of a variable to affect the representation of other parts of the program where the variable is used.

Multi-head attention allows for the different heads to learn different aspects of the data. For our problem of program equivalence, the model is trained to produce correct rewrite rules and hence all of the learnable functions are learning representations useful for this task. To illustrate, one of the heads may tend to build a representation for addition and subtraction of vectors, while another head might build a representation for multiplication and division of scalars. What the heads learn is not constrained by the architecture, so the true meaning of the heads at each layer is not easily decipherable, but multi-head attention has been shown by others and by our own ablation studies to be valuable.

Figure~\ref{fig:Transformer} shows the architecture with multiple layers for both the encoder on the left (which encodes the input programs to an intermediate representation -- an embedding) and the decoder on the right (which decodes the representation to produce the rewrite rule output). The multiple layer encoder is using self-attention in which the information processed by a given layer is provided by the layer below. A similar situation holds for the multiple layer decoder, however the $H$ connection from the encoder layers to the decoder layers allows the decoder to process information from both the decoder and encoder.
The transformer model we employ also includes residual and feed-forward sublayers and a full description of the interactions within the model can be read in the work by Vaswani \etal \cite{vaswani2017attention}. We used the Adam optimizer \cite{kingma2015adam} to adjust the weights in the network based on the loss function between the target token expected in the training sample and the result produced by the current network.

The intermediate representation $H$ encodes the complex representations for each token of the input, in our case a pair of programs. The decoder model will generate a first output token based on $H$ and then generate subsequent tokens based on $H$ and the previously output tokens. The Softmax function normalizes the output of the final decoder layer to a probability distribution:
\[
\text{Softmax}(z_{i}) = \frac{e^{z_i}}{\sum_{j=1}^K e^{z_j}} \texttt{ for } \textbf{z} = (z_1,...,z_K) \in \mathbb{R}^K
\]
The effect of the Softmax layer is to create an output that can be interpreted as representing the probability that a given token is correct, given the training the model has been exposed to. As the output is generated, when the tokens with the highest Softmax values are selected we create a rewrite rule which represents the most likely next edge in our path through the program space from ProgA to ProgB.

\subsection{Beam Search to Explore Multiple Possible Proofs}
\label{sec:proofsearch}

A typical approach when using sequence-to-sequence models is called ``beam search'', it is the process of asking the deep learning model to produce multiple answers for the same question. As the network produces likely token selections, the Softmax layer shown in Figure~\ref{fig:Transformer} is effectively producing probabilities for each token. Using these probabilities we can produce multi-token outputs sorted in order based on the total probability the model assigns to the sequences.

In \DLTOOL, we use this beam search to enumerate the possible next rewrite rule to apply.
Each proposal is then checked for legality (whether the proposed rewrite rule can indeed be applied at the given program location) and novelty (the program resulting from the rule application does not match a previously seen program for this search). Figure~\ref{fig:ProofSearch} diagrams the system which receives ProgA and ProgB and generates a verifiable proof of equivalence. A rewrite rule includes the statement number, rule name, possible statement node identifiers, and possible variable identifiers. The validity of a rule is checked by first verifying the statement number exists in ProgA. Then, if the rule does not require a node ID (SwapPrev or DeleteStm) we check if the liveness rules are followed - for SwapPrev neither statement may make use of the variable assigned by the other statement, and for DeleteStm the variable must not be an output variable and must not be used in subsequent statements. For rules requiring a NodeID the operators or variables are checked for legality of the operation being applied along with the possible variable ID produced for use by the rule.

In addition to the rewrite rule beam, \DLTOOL uses a second type of beam during the proof search.
The idea is to feed the network again based on the result of the application of the previously suggested rewrite rules.
We denote the number of enumerated rewrite rules $b$.
As the search advances, the $b$ outputs from the neural network may all lead to potential intermediate programs from which a search can continue. After having checked for legality, we limit this potential exponential growth in the search to at most configurable $i$ intermediate programs which may be explored at a given proof step.

Consider a search where we set $b$ to 3 and $i$ to 2, as diagrammed in Figure~\ref{fig:ProofSearch}. When a transformation search between 2 programs is attempted, at first there is only 1 sample (the original 2 programs to prove equivalent) fed into the transformer model which will propose 3 rewrite rules. Perhaps 2 are legal rewrite rules; both of these are checked for equivalence to the ProgB goal and assuming there is no match both will be fed into the transformer model on the next step. This will produce 6 proposed rewrite rules. If a rewrite rule would create an intermediate program that is already being searched (for example, commuting the same node twice in a row) then the search process will not create the duplicate intermediate program. In the figure, rewrite rules which are illegal or create duplicate search programs are marked with a red X. The search routine will select the most likely proposed rule for each ProgInt/ProgB pair if it legally creates a novel intermediate program.

As diagrammed, the ProgInt$_{11}$/ProgB pair produces a legal novel program which is used for the next step, but the ProgInt$_{12}$/ProgB pair's 1st proposal is not usable. Since the 2nd proposal from the ProgInt$_{11}$/ProgB pair is also not usable, the legal 2nd proposal from the ProgInt$_{12}$/ProgB pair is used. Our search will first try to use the 1st proposed rule from each ProgInt/ProgB pair, then the 2nd, and so on until the next $i$ intermediate programs are created. We will limit the intermediate programs that feed into the transformer to $i$ as the search is continued up to the rewrite rule sequence step limit $l$ (such as 25 steps). In rare cases, none of the proposed rewrite rules for any of the intermediate programs will produce a legal novel intermediate program and the search will terminate before the step limit. In our example, the 2nd rewrite rule proposed from the model when given ProgInt$_{12}$/ProgB to the transformer rewrites ProgInt$_{12}$ into the lexical equivalent of ProgB, and hence, a 2 step proof has been found. ProgA is transformed into ProgB by applying Rewrite Rule$_{13}$ and then Rewrite Rule$_{21}$.

\subsection{Naming Conventions Used}

Throughout this paper we make reference to various data sets and and configuration parameters. As an aid to understanding our system and experiments, we summarize them in Table~\ref{tab:datasets} with a short description. Our $Synth$ datasets are generated algorithmically as described in section~\ref{sec:synthdata}. Our $Optim$ datasets are abstractions from GitHub source C code as described in Section~\ref{sec:githubdata}.

\begin{table}
  \caption{Variable and dataset names\vspace{-.1cm} }

  \label{tab:datasets}
  \footnotesize
  \setlength\tabcolsep{3pt}
  \centering
  \begin{tabular}{@{}lrr@{}}
    \toprule
    Variable or set description & & Name \\
    \cmidrule{1-1} \cmidrule{3-3}
    Beam search width produced by neural network & & $b$ \\
    Number of intermediate programs during search & & $i$ \\
    Rewrite rule step limit for equivalence search & & $l$ \\
    Initial model trained only with supervised data & & $M_1$ \\
    Models trained with unsupervised data & & $M_{2-7}$ \\
    Baseline model fully trained with supervised data & & $Q$ \\
    Synthetically generated program pairs & & $Synth^{+rules}$ \\
    \hspace{0.2cm}(includes known rewrite rules between pairs) & & \\
    Synthetically generated program pairs & & $Synth_{val}$ \\
    \hspace{0.2cm}for validating models (no rules included) & & \\
    Synthetically generated program pairs & & $Synth_{train}$ \\
    \hspace{0.2cm}for training models (no rules included) & & \\
    Synthetically generated program pairs & & $Synth_{test}$ \\
    \hspace{0.2cm}for final testing of models (no rules included) & & \\
    Code optimization pairs from human-written code & & $Optim_{val}$ \\
    \hspace{0.2cm}for validating models (no rules included) & & \\
    Code optimization pairs from human-written code & & $Optim_{train}$ \\
    \hspace{0.2cm}for training models (no rules included) & & \\
    Code optimization pairs from human-written code & & $Optim_{test}$ \\
    \hspace{0.2cm}for final testing of models (no rules included) & & \\
    Program pair abstractions from human-written code & & $Human$ \\
    Initial training set derived from $Synth^{+rules}$ & & $T_1$ \\
    Incremental training set derived from proof attempts & & $T_{2-7}$ \\
    on $Synth_{train}$ and $Optim_{train}$ & & \\
    \bottomrule
  \end{tabular}
\end{table}

\subsection{Training Process}
\label{sec:s4}

\begin{figure}[h!tb]
\vspace{-.2cm}
\centering\includegraphics[width=9cm]{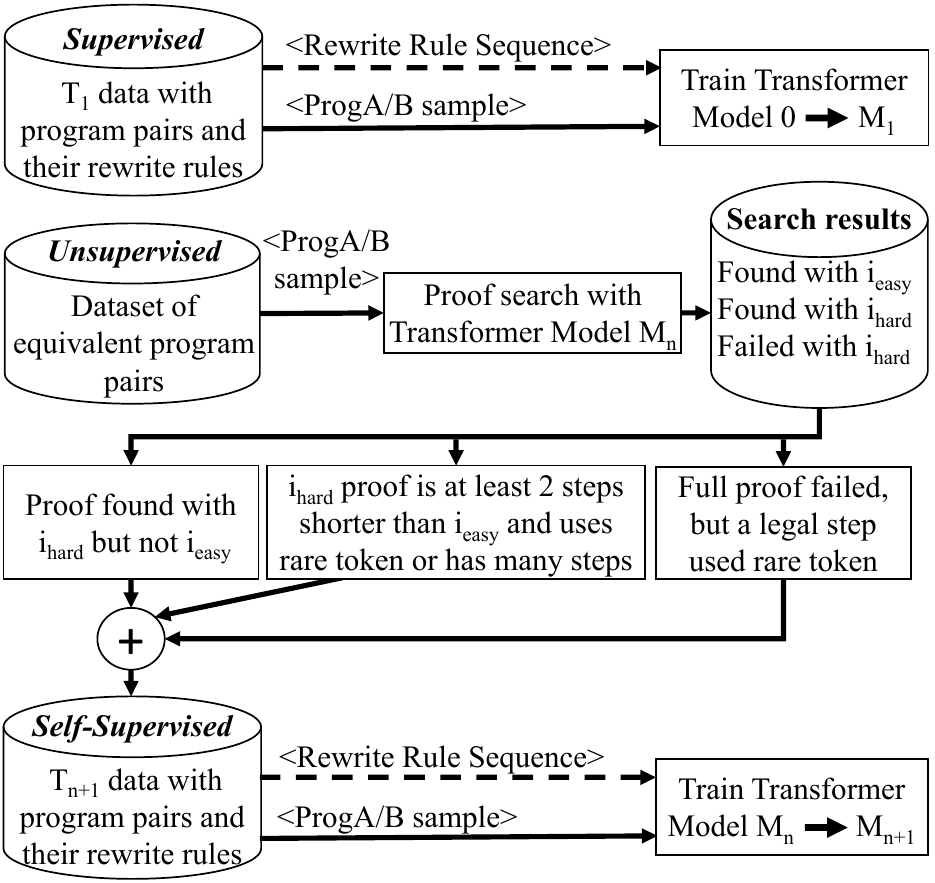}
\caption{Self-Supervised Sample Selection: proof attempts with $i=i_{easy}$ and $i_{hard}$ intermediate programs are used to incrementally train the model.}
\vspace{-.2cm}
\label{fig:Training}
\end{figure}

A key novelty of \DLTOOL lies is the way we train the neural network.
We devise an original training process which is dedicated to the challenges of synthesizing equivalence proofs.
This training process involves three kinds of training which we now present.

\subsubsection{Initial Supervised Training}
\label{sec:supervised}
The typical technique for training sequence-to-sequence models involves providing supervised training samples which provide the target output paired with sample input.
\DLTOOL performs initial training using program pairs for which a rewrite rule sequence is known between the two programs ProgA and ProgB.
The details of how to obtain such a supervised dataset are discussed in
Section~\ref{sec:datagen}.
We refer to this initial supervised dataset of program pairs and their rewrite rules as $T_1$, and the we refer to the initial model trained with it as $M_1$.

\subsubsection{Incremental training with challenging proofs}
\label{sec:incrementallearning}

After we have an initially trained model, we can use it to attempt new proofs. Because a proposed rewrite sequence between 2 programs can be checked automatically, we can automatically verify the generated outputs and use them to further optimize the model.
At this point, to further optimize the model, we don't need to generate program pair inputs with rewrite rule outputs in a supervised manner, but only program pair inputs. In other words, this can be considered as self-supervision since the labeling is automated.

To be effective, the core challenge becomes to effectively select new samples.
Hence, we term this technique \emph{self-supervised sample selection}, and since our framework uses this technique on the problem of program equivalence, we name our framework \DLTOOL.

First, we need a dataset (or a data sample generator) of known equivalent programs which does not include the rewrite rule steps.
This dataset is unsupervised, as the pairs do not have target rewrite rules associated with them. Relative to the samples in $T_1$, the looser restrictions on the unsupervised dataset allows \DLTOOL to generalize to program distributions and proof distributions which may be different than those found in $T_1$.

Figure~\ref{fig:Training} gives a complete overview of our training process for the transformer models.
As we show in Section~\ref{sec:supervised}, we start with supervised training and produce model $M_1$ using program pairs and their expected rules.
Now we can use model $M_1$ to attempt proofs with known equivalent programs in the unsupervised dataset.

Our key idea is to focus on challenging problems for the current model. Given an intermediate program limit, we can say that the model can 'easily' prove equivalence if a small number of intermediate programs are available at each search step (such as 1 or 2). In such a case, the most likely rewrite rules proposed by the model are checked for equivalence up to a given proof step limit, and if the proof is found, then the model is already well trained for the given problem. Our variable representing the number of intermediate programs searched at each proof step is $i$ and the number of steps to search is $l$ steps. To check for 'easy' proofs, we search for a proof with $i=i_{easy}$ where $i_{easy}$ is a small number. To check for 'challenging' proofs, we set $i=i_{hard}$ where $i_{hard}$ is a larger number (10 or more), allowing the proof search to explore more possible rewrite rule paths which are considered less likely to be correct by the current model. We attempt to prove each known-equivalent pair with $i=i_{easy}$ and also $i=i_{hard}$.
We define challenging proofs as those found with $i_{hard}$ but not $i_{easy}$.
When proofs are attempted with model $M_n$, the next training dataset ($T_{n+1}$) includes samples with proof steps found with $i_{hard}$ but not $i_{easy}$;
thus, the model is more likely to propose similar steps in the future.

In addition to including challenging proofs, if $i_{hard}$ found a proof at least 2 steps shorter than the $i_{easy}$ proof we include that proof given certain conditions. We set the probability of including such proofs based on length in order to bias the self-supervised samples to solve complex proofs. Also, based on distributions discussed in Section~\ref{sec:synthdata}, we include the $i_{hard}$ proofs when they include rare output tokens such as referring to higher statement numbers, or deeper expressions nodes, or rare rewrite rule names.


After creating our initial training samples in $T_1$, we train on various model hyperparameter options and use validation test sets to determine the model best suited for continued training. During incremental training, the model size parameters (number of layers, \etc) are constant but we train with variations on initial learning rate and decay rate. We then select the best model to continue training using the same validation sets. These validation sets help prevent catastrophic forgetting  \cite{goodfellow13}. But additionally, if a model becomes weaker at solving certain problems then problems similar to those will get selected in the next iteration of the training, again reducing catastrophic forgetting.

Boosting methods in which models weak in one part of the problem distribution are boosted by other models have been shown to reduce risk of overfitting \cite{Dietterich00} and we anticipate that our methodology for incremental training based on challenging proofs will similarly resist overfitting.

%
%
%
%
%
%
%
%
%
%
%
%
%
%
%
%
%
%

\subsubsection{Incremental training with rare tokens}
\label{sec:hindsight}

If some input or output tokens are not yet well understood by a given model $M_n$, it is because the training datasets so far do not have sufficient samples demonstrating the use of those tokens. To overcome this problem, we propose another kind of incremental training based on rare tokens.
The core idea is to oversample those proofs and rewrite rules that involve rare tokens.
Even when a full proof search fails, when a rare output token is used legally by a single rewrite rule step in the proof, we keep it in the training dataset.
This type of training improves the model representation for the rare token and the situations in which it should be applied and is based on the hindsight experience replay concept \cite{andrychowicz17}.

Consider again Figure~\ref{fig:ProofSearch} and a case where a required rewrite rule to prove ProgA equal to ProgB was, for example, \texttt{stm2 Commute Nlrll}. The node \texttt{Nlrll} is an example of a node ID which specifies the 5th level of the AST for statement 2. If there haven't been sufficient training samples with \texttt{Nlrll} then the model may not have a good internal representation for when the node should be produced by the final Softmax layer of the transformer model and the proof might fail. However, if, for example, $RewriteRule_{21}$ was \texttt{stm2 AssociativeLeft Nlrll} and this was a legal application of \texttt{AssociativeLeft} then the pair with $ProgInt_{11}$ and $ProgInt_{21}$ can be proven equal by applying $RewriteRule_{21}$. If \texttt{Nlrll} is a rare token, this sample can be included in the next training dataset and this will improve the model's representation for this rare token in order to improve use of this token after incremental training.


\section{Experimentation}
\label{sec:experimentation}

We now describe our experiments to assess \DLTOOL.
We start by carefully devising two datasets for training and evaluating our system.

\subsection{Dataset Generation}
\label{sec:datagen}

We devise two separate datasets with different properties. First, we wish to evaluate our algorithm broadly on code optimizations of human-written programs from open source C functions found on GitHub, we call this dataset $Optim$.
Second, we develop a process to create synthetic program pairs based on applying rewrite rules, we call this dataset $Synth$.

\subsubsection{Equivalent program pairs from GitHub}
\label{sec:githubdata}

We want to have  a dataset representative of developer code with straight-line programs matching our grammar.
For this, we use an existing dataset of C programs mined from GitHub suitable for machine learning \cite{chen2019sequencer}.

We process these C functions to find sequences of assign statements that correspond to our straight-line program grammar.
We search for C snippets of mathematical computations, with at least 2 assignments, at least one multiply or divide statement, and require at least 1 temporary variable is used in an output variable. To create program abstractions (a process similar to function outlining \cite{Zhao07}), we collapse complex data structure accesses into an intermediate variable.
For example, C code of the form \texttt{delta = ca->tcp\_cwnd - cwnd ; max\_cnt = cwnd / delta ;} will be abstracted to \texttt{ t1 = i1 - i2 ; o1 = i2 / t1 ;}. Two complete examples of abstracted programs are shown in subfigures~\ref{fig:EquivExample}(b) and ~\ref{fig:EquivExample}(e).

\begin{algorithm}[h!tb]
\DontPrintSemicolon
\SetAlgoLined
\SetKwInOut{Input}{Input}\SetKwInOut{Output}{Output}
\Input{Tokenized C Functions from GitHub: $F$}
\Output{Prefix encoded equivalent pairs created using code optimizations: $Optim$}
\BlankLine

$Optim \leftarrow \emptyset$ \hspace{0.2cm}\{Compiler equivalence program pairs\}

$S \leftarrow $FindSourcePrograms$(F)$ \hspace{0.2cm}\{Possible programs\}

\ForEach{$s$ in $S$}{
 $p \leftarrow $Encode$(s)$ \{Encode into prefix format\}

 $p_{cse} \leftarrow $CommonSubexprElimination$(p)$

 $p_{str} \leftarrow $StrengthReduction$(p_{cse})$

 $p_{reuse} \leftarrow $VariableReuse$(p_{str})$

 $p_{rules} \leftarrow $Rules$(p_{reuse})$ \{Probabilistic application\}

 \If{CheckLimits$(p,p_{cse},p_{str},p_{reuse},p_{rules})$}{
   $Optim \leftarrow Optim + $MixPairs$(p,p_{cse},p_{str},p_{reuse},p_{rules})$
  }
}

\caption{GenerateKnownEqual}
\label{alg:GenerateKnownEqual}
\end{algorithm}

Algorithm~\ref{alg:GenerateKnownEqual} provides an overview of the process we use. After finding C source code GitHub programs, we perform 3 high-level compilation steps: common subexpression elimination, strength reduction, and variable reuse.
For training sample generation, Encode will transform C code into the prefix encoding of the AST. Encode will randomly reassign scalar variable IDs to temporary variables with each iteration of the \texttt{foreach} loop; so \texttt{t1} may be assigned to \texttt{s03} for one program and \texttt{s25} for another program. The goal of the random assignment by Encode is to help with generalization.
The high-level compilation steps utilize many of the 23 rewrite rules shown in Table~\ref{tab:RulePct}, but in order to ensure all rewrite rules are represented in $Optim$, we call the Rules function on $p_{reuse}$ (the encoded program after all compilation steps) which may apply one or more rewrite rules to create $p_{rules}$. The Rules function is used heavily in Section~\ref{sec:synthdata} and is discussed further there.
The CheckLimits function ensures that our samples meet the model limits\footnote{We limit each  program to 20 statements, at most 100 AST  nodes, at most 30 scalar variables, an expression depth of 5 levels of parenthesis (expression AST depth of 6), and at most 2 outputs. }.


Using these rules, the dataset $Optim$ derived from C functions from Github eventually contains 49,664 unique known equivalent program pairs for our experimentation. Note this approach does fully preserve the shape of the AST of the original GitHub C code snippet. It does not however preserve names, since we rename all distinct variables/array accesses to a unique and canonical name for that program. For example, \texttt{particles[i].y\_pos} becomes \texttt{i1} in the generated program, while \texttt{y} is \texttt{i1} in Fig.~\ref{fig:EquivExample}.

\subsubsection{Synthetic equivalent program pairs}
\label{sec:synthdata}

As we discuss in Section~\ref{sec:s4}, we need to create an initial training set with broad distribution on the input and output tokens necessary for our problem of proving programs equivalent.
We create legal input programs by probabilistically applying production rules from a grammar which defines our target program space. This approach allows us to create arbitrarily large amounts of training data.

\begin{algorithm}[h!tb]
\DontPrintSemicolon
\SetAlgoLined
\SetKwInOut{Input}{Input}\SetKwInOut{Output}{Output}
\Input{Probabilistic grammar: $G$, Number of samples desired: $n$}
\Output{Prefix encoded equivalent program pairs with rules: $Synth^{+rules}$}
\BlankLine

$Synth^{+rules} \leftarrow \emptyset$ \hspace{0.2cm}\{Rewrite rule equivalence program pairs\}

\While{samples in R < n}{
  $p_A \leftarrow $GenerateProgA$(G)$ \{Probabilistic production rule generation\}

  $p_B \leftarrow $Rules$($Rules$($Rules$(p_A)))$ \{3 passes over $p_A$ with probabilistic application of rewrite rules\}

  \If{CheckLimits$(p_A,p_B)$}{
    $Synth^{+rules} \leftarrow Synth^{+rules} + (p_A,p_B,rules)$
  }
}

\caption{GenerateRewrites}
\label{alg:GenerateRewrites}
\end{algorithm}

\begin{figure}[h!tb]
\vspace{-.4cm}
\centering\includegraphics[width=9cm]{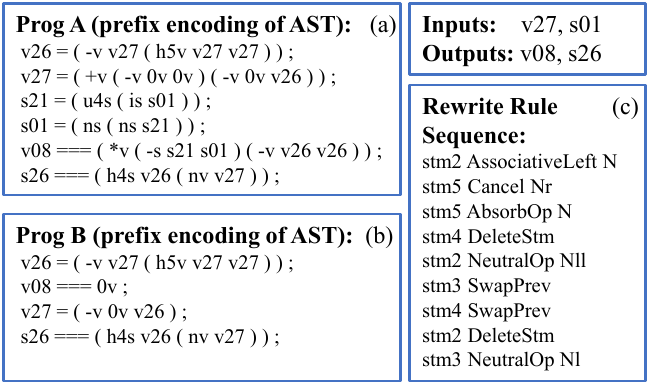}
\caption{Equivalence proven between 2 multi-statement programs generated synthetically. The equivalence proof is 9 steps long involving expression and statement rules.}
\vspace{-.1cm}
\label{fig:SynthExample}
\end{figure}

Algorithm~\ref{alg:GenerateRewrites} shows the synthesis algorithm.
Our program grammar defines a program as made up of a series of assign statements which assign scalar and vector variables to complex mathematical expressions.
Our program generation process starts by creating assignment statements for the output variable(s). Then a subset of the variables used in the expression may have earlier assign statements added. This process continues adding assign statements randomly to the beginning of the program.

Variables which are used but never assigned are considered program inputs. For example, in the program ``\texttt{c = b + a; d = c * a + b;}''; a and b are inputs, d is an output, and c is a temporary variable. Subfigure~\ref{fig:SynthExample}(a) shows an example ProgA generated using our algorithm. It includes 6 statements and produces one vector output v08 and one scalar output s26 identified with \texttt{===} tokens.

\emph{Rules}
In order to create training samples with known paths between equivalent programs, after creating a program we randomly apply legal rewrite rules to the start program. For example, \autoref{fig:SynthExample}(c) shows the rewrite rules randomly selected which transform ProgA in subfigure (a) into ProgB in subfigure (b).

\begin{figure}
\centering\includegraphics[width=9cm]{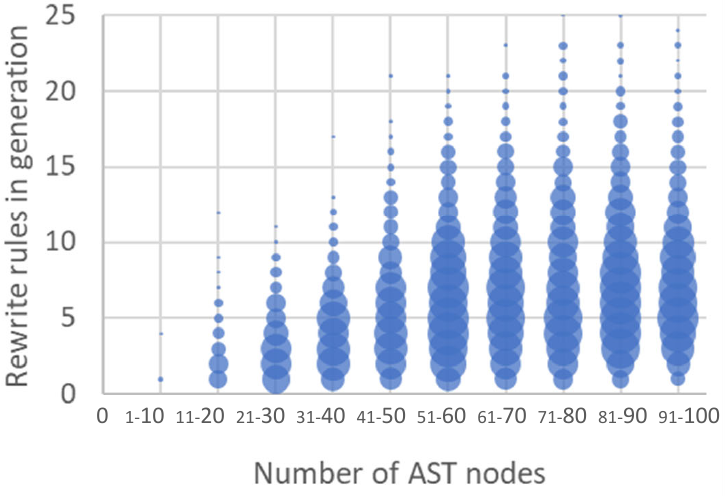}
\caption{Distribution of proof length for synthetic program equivalence dataset $Synth^{+rules}$.}
\vspace{-.4cm}
\label{fig:SynthDataset}
\end{figure}

\emph{Synthetic distribution}
Figure~\ref{fig:SynthDataset} diagrams the distribution of samples generated by Algorithm~\ref{alg:GenerateRewrites} with a plot of the number of AST nodes in ProgA and the number of rewrite rules used to generate ProgB in the sample. When GenerateProgA generates a program with more AST nodes, more rewrite rules are found by invoking of Rules function. We limit the number of AST nodes to 100, as shown in the distribution, but the number of rewrite rule steps is not strictly limited and we have some cases with over 40 steps between ProgA and ProgB (example long proofs can be found in our GitHub for this paper \cite{KommruschS4Eq21}).

Certain rewrite rules, such as \texttt{Commute}, tend to be applicable to many nodes in a program AST, while others, such as \texttt{FactorLeft}, require rarer patterns in order to be legally applied.
We adjust the likelihood of rewrite rules being applied to help balance the likelihood a proof will use any given rewrite rule. All 23 rules shown in Table~\ref{tab:RulePct} will occur in our synthetic dataset. Because the \texttt{Commute} rule can be applied to a large number of operators in our AST, we limit the likelihood it will be applied to about 9\% per location with the result that 60.2\% of the proofs use it for generating ProgB. Conversely, given our random program generation process, \texttt{FactorLeft} and \texttt{FactorRight} are not so likely to be applicable, so we bias the application so that about 65\% of the AST nodes which would allow these rules have them applied which results in 7.1\% and 7.2\% of proofs using these rules.

For algorithm~\ref{alg:GenerateRewrites} the probabilistic grammar expansion used in GenerateProgA is tuned so as to create a range of program sizes. We skew the generation to prefer creation of programs with large AST node counts which have either many statements or deep expressions and rely on CheckLimits to insure programs outside our transformer model ranges are pruned out. For the $T_1$ initial training dataset, we create 150,000 equivalent program pairs; given that the average pair takes multiple rewrite rules to transform, we create 640,000 rewrite rule steps from these pairs for training the $M_1$ model.

Finally we note our set is made of about 90\% of programs where at least one live-in symbol is used in computing some output, ensuring a high (but not exclusive) representation of "useful" programs where dead code elimination cannot trivially solve their equivalence. Dependencies between statements are produced during program generation; a variable previously written in a prior statement may be used in the currently synthesized expression statement.

\subsection{Experimental setup}
\label{sec:exptrain}

Recall from Section~\ref{sec:s4} that the model initially trained on synthetic programs with target rewrite rules is labeled $M_1$. We train our initial model $M_1$ for 100,000 steps, which is 5 epochs of our 640,000 sample $T_1$ dataset (after dividing by our effective batch size of 32). Figure~\ref{fig:ProofSearch} diagrams the proof search process which allows unsupervised samples without known proof steps to be labeled with a valid rewrite rule sequence. As per Figure~\ref{fig:Training}, $M_7$ has gone through 6 iterations of incremental training beyond the initial $M_1$ model. In order to validate partially trained models during training, we create $Synth_{val}$ and $Optim_{val}$, which are each 1,000 equivalent program pairs sampled from the $Synth$ and $Optim$ datasets. The success of the intermediate models on proving equivalences in $Synth_{val}$ and $Optim_{val}$ is used to select the model to use in the next step of incremental training. For a more complete study of our final trained model $M_7$ we create $Synth_{test}$ and $Optim_{test}$ each with 10,000 equivalent program pairs for evaluation.

\begin{figure}
\centering\includegraphics[width=7cm]{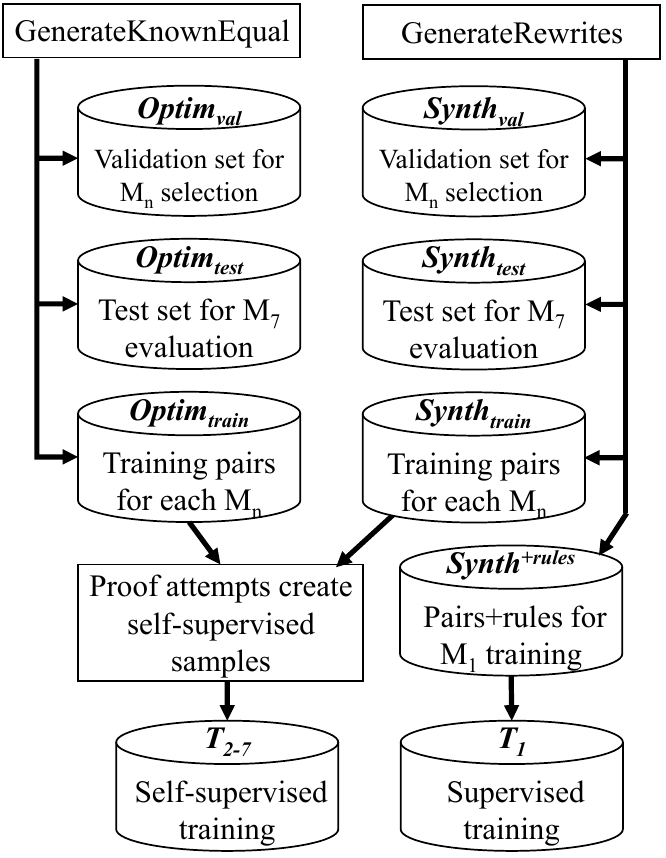}
\caption{Generation of program equivalence datasets used for model training and evaluation.}
\vspace{-.4cm}
\label{fig:DataGen}
\end{figure}

Figure~\ref{fig:DataGen} diagrams how the algorithms introduced in Sections~\ref{sec:githubdata} and ~\ref{sec:synthdata} are used along with self-supervised sample selection introduced in Section~\ref{sec:s4} to create training samples $T_{1-7}$ to train our models. As we generate the initial supervised training dataset $T_1$ we also generate the 2 validation datasets and 2 test datasets and we remove any duplicate program pairs. As we generate $T_{2-7}$ we also ensure duplicates with the validation and test datasets are removed. \revision{Because of our generation method, the distribution of training, validation, and testing datasets are very similar - i.e. the average program size, average program AST depth, etc., will all be nearly the same for the 3 groups. However, in Section~\ref{sec:generalize}, we study 2 groups of program pairs which have distributions outside of these 3 sets - synthetic programs of longer length and program pairs created by pairing GitHub samples with each other.}

For self-supervised sample selection, after training $M_1$ with $T_1$ we want to create training data which insures we continue improving performance on the synthetic dataset $Synth$ but also learn to solve equivalent programs in $Optim$. Referring to Figure~\ref{fig:Training}, we create the unsupervised set of program pairs at each iterative learning step by selecting 60,000 equivalent pairs from $Synth$ and 40,000 equivalent pairs from $Optim$.

In order to attempt to prove equal the unsupervised program pairs, we chose $i_{easy}=2$ for the ``easy'' beam width as a beam width of only 1 can fail with a single misstep and we wanted to allow recovery from that case. Due to machine time constraints, we chose $i_{hard}=20$ for the ``hard'' beam width and $s=25$ for the maximum number of proof steps to search. From this data, we create $T_n$ for use in training model $M_n$.

During incremental training, we train 4 model versions (we vary the learning rates) for 50,000 steps with the new $T_n$ training dataset. Similar to early stopping techniques \cite{prechelt1998early}, we use our validation datasets to select the best performing model during incremental training. We save the model on disk every 10,000 training steps and select $M_n$ as the model with the highest total proofs found in the 2,000 test samples when $Synth_{val}$ and $Optim_{val}$ are combined.
Our final iteration is selected when both $Synth_{val}$ and $Optim_{val}$ have improved performance by 1\% or less when training the next model. For our experiments, this is $M_7$.

We perform all our experiments on systems with 12 Intel(R) Xeon(R) CPU E5-1650 v4 @ 3.60GHz CPU cores and NVIDIA GeForce GTX 1060 6GB for GPU compute. Our model is based on the transformer model implemented in OpenNMT-py.

\subsection{Research Questions}
\label{sec:expprotocol}

In this section, we describe our research questions for \DLTOOL and the protocol methodologies for evaluating them.

Our research questions are:
\begin{itemize}
\item RQ1: How effective is \DLTOOL on the synthetic test dataset?
\item RQ2: How useful is incremental learning with self-supervised sample selection in improving \DLTOOL's model?
\item RQ3: To what extent does our model generalize outside the training data?
\end{itemize}

\subsubsection{Methodology for RQ1}
\label{sec:RQ1method}

To evaluate the effectiveness of \DLTOOL on finding rewrite rule sequences for the program pairs selected from the synthetic program dataset $Synth$, we analyze the distribution of programs and rewrite rules included in $Synth$ and present data on the success rate for proving various subsets of the test dataset $Synth_{test}$.
Our goal is to understand if our system is unable to perform well on any of our selected subsets of $Synth_{test}$ (\ie solve at less than a 90\% success rate).

\subsubsection{Methodology for RQ2}
\label{sec:RQ2method}

Our incremental training approach is able to learn to solve proofs for which the supervised proof sequence is not provided during sample generation.
To determine how well self-supervised sample selection improves the quality of the \DLTOOL model, we study the 10,000 sample $Optim_{test}$ dataset drawn from $Optim$ alongside our rewrite rule test dataset $Synth_{test}$.

In addition to our golden model $M_7$, which is trained using self-supervised sample selection, we create a model for comparison called $Q$ which is trained in a more traditional way. $Q$ is trained for the same number of training steps as $M_7$ but it continues training on the $T_1$ dataset from the $M_1$ model in a supervised manner. To align with our $M_n$ training protocol, we train multiple models from $M_1$ with varying learning rates and select the strongest model using the $Synth_{val}$ and $Optim_{val}$ datasets. We will evaluate the ability of our models to prove the known equivalent program pairs in $Optim_{test}$ in order to determine if self-supervised sample selection adds value to our model.

\subsubsection{Methodology for RQ3}
\label{sec:RQ3method}

We explore the ability of \DLTOOL to generalize using 2 different methods.
The first method focuses on using the sample generation algorithm~\ref{alg:GenerateRewrites} for $Synth$ with different hyperparameters to create programs outside of the training distribution. The second method relates to actual use cases for program equivalence and focuses on finding equivalent programs from within different GitHub C functions. This search could be viewed as a demonstration of the system to find semantic equivalence for a variety of uses, such as identifying opportunities for library calls \cite{Mora18}, or grouping student submissions into semantically equal groups for grading \cite{Clune20}, \etc

Regarding the first method, we use algorithm~\ref{alg:GenerateRewrites} to create programs with exactly 3 outputs and 101 to 120 AST nodes. Recall that in training we limit $Synth$ and $Optim$ to include only 1 or 2 output programs with up to 100 AST nodes. We test our golden model on this dataset to determine if it has overfit the training data or if can solve problems from outside the training distribution.

Regarding the second method, the FindSourcePrograms routine in algorithm~\ref{alg:GenerateKnownEqual} finds 13,215 unique multi-statement program blocks from GitHub. No pair of these programs are lexically equivalent and hence at least one rewrite rule step will be required for \DLTOOL to prove equivalence. As we are interested in comparing the more complex programs in this set, we select only programs with at least 30 AST nodes resulting in a set of 4,600 programs. We then group the programs and test all pairs of programs which have the same number of inputs, outputs, and functions. This results in 152,874 unique pairs of programs to check for equivalence. Since we search in both directions for each pair, our full GitHub test set $Human$ has 305,748 program pairs to attempt equivalence proofs on.

\subsection{Experimental Results}
\label{sec:expresults}

\subsubsection{RQ1: Effectiveness on Synthetic Test Dataset}
\label{sec:expsynth}

In this research question, we aim to show the breadth of program pairs and rewrite rule sequences in our synthetic dataset and to demonstrate the effectiveness of our $M_7$ golden model.
Table~\ref{tab:categories} details the performance of a variety of subsets of our 10,000 pair synthetic test dataset $Synth_{test}$.
Each cell in the table gives the passing rate and sample counts for each subset.
Here the passing rate means the percentage of program pairs proven equivalent with an appropriate sequence of rewrite rules (recall that all program pairs in $Synth$ are equivalent by construction).
The first data row gives the effectiveness over the whole dataset.
The other rows in the table provide data on subsets of the sample based on the rewrite rules used to generate the ProgB in the sample given the synthetically generated ProgB. A full description of the rows is given in Section~\ref{sec:expprotocol}. Orthogonal to the rewrite rules used to generate ProgB, the columns explore subsets of interest for the original ProgA in the sample. The first column provides data for all samples which conform to the rewrite rule subset given by the rows. The 2nd column shows results for the 6,390 samples that used at least 3 functions in ProgA. The 3rd column shows results for the 3,340 samples where ProgA starts with an expression of depth 4-6 (note that a NodeID at depth 5 is used in 533 samples given any ProgA but only 506 samples when ProgA started with a deep expression; this is due to some rewrite rules, such as \texttt{MultOne}, adding depth to the original ProgA). The final column provides data on the larger programs from the dataset (and the size corresponds to the program sizes considered in RQ3).

In the upper left data cell in Table~\ref{tab:categories}, we find that of the 10,000 samples in $Synth_{test}$, $M_7$ was able to find a rewrite rule sequence from ProgA to ProgB for 98\% of them, which is arguably very high. We see that some subsets of $Synth_{test}$ performed better than this overall result, such as cases where ProgA to ProgB proofs contain 1-10 rewrite rule steps (99\% effectiveness). Other subsets performed worse, such as proving samples where a depth 5 NodeID was use to create ProgB. In general we see that program pairs generated with shorter rewrite rule sequences are more easily proven equal than longer ones, corresponding to the intuition that shorter proofs are easier to find. In the table, there are 5 subsets tied for the poorest result of 92\% success: all 4 subsets with NodeID at depth 5, and cases where ProgA has a deep expression and generating ProgB used over 10 rewrite rule steps. These results show that \DLTOOL is challenged by deeply nested expressions and long proofs, however it still achieves over 90\% success.

\begin{table}
  \caption{Success rate for subsets of test dataset $Synth_{test}$. Each entry includes tuned passing percentage and sample count within $Synth_{test}$.\vspace{-.1cm} }

  \label{tab:categories}
  \scriptsize
  \setlength\tabcolsep{3pt}
  \centering
  \begin{tabular}{@{}lrrrrrrrr@{}}
    \toprule
     & & & & & & Maximum & & \\
     & & & & Functions & & Expression & & Nodes \\
    Rewrite Rules & & ALL & & 3 or more & & Depth 4-6 & & 30-100 \\
    \cmidrule{1-1} \cmidrule{3-3} \cmidrule{5-5} \cmidrule{7-7} \cmidrule{9-9}
    Whole dataset & & 98\%(10000) & & 98\%(6390) & & 97\%(3340) & & 98\%(9470) \\
    Rename & & 97\%(552) & & 97\%(354) & & 95\%(74) & & 97\%(546) \\
    Newtmp & & 94\%(844) & & 94\%(591) & & 93\%(477) & & 94\%(828) \\
    DistributeLeft & & 96\%(2273) & & 95\%(1650) & & 95\%(1524) & & 96\%(2214) \\
    No statement rules & & 99\%(4923) & & 98\%(3208) & & 98\%(2312) & & 99\%(4524) \\
    NodeID at depth 5 & & 92\%(533) & & 92\%(429) & & 92\%(506) & & 92\%(522) \\
    Rewrite steps 1-10 & & 99\%(8253) & & 99\%(5066) & & 99\%(2269) & & 99\%(7725) \\
    Rewrite steps 11+ & & 94\%(1747) & & 93\%(1324) & & 92\%(1071) & & 94\%(1745) \\
    \bottomrule
  \end{tabular}
\end{table}

Table~\ref{tab:categories} also shows that \DLTOOL handles rare tokens well.
The output token least represented in our 640,000 samples in $T_1$ is \texttt{Nrlrr}, used in only 278 samples. \texttt{Nrlrr} is one of 16 tokens used to indicate that a rule should be applied to a depth 5 node; all 15 of the other such tokens make up the 15 other least commonly used tokens in the $T_1$ output group. During self-supervised training, our hindsight methodology helps compensate for this rarity to increase the number of such samples in $T_{2-7}$.
As we show with the ``NodeID at depth 5'' row in Table~\ref{tab:categories}, our fully trained model has learned to use these tokens effectively.

Our synthetic ProgA algorithm given in algorithm~\ref{alg:GenerateRewrites} is designed to insure a relatively balanced use of input tokens in $T_1$ for training.
This works well, because the least used input token is one of the 5 tokens representing functions which receive 2 scalars and produce a vector output (\texttt{f4v}) with 43,458 of the 640,000 samples using it. Table~\ref{tab:categories} shows that those programs using at least 3 functions are proven well by \DLTOOL.

\begin{mdframed}
Answer to RQ1: \DLTOOL is effective on the $Synth$ dataset, achieving over 90\% successful proofs on all subsets analyzed. \DLTOOL handles well both rare input tokens used in the programs to be analyzed (such as function names) and rare output tokens (such as depth 5 NodeIDs).
\end{mdframed}

\subsubsection{RQ2: Incremental Training Benefit}
\label{sec:exps4}

\begin{figure}
\vspace{-.2cm}
\centering\includegraphics[width=9cm]{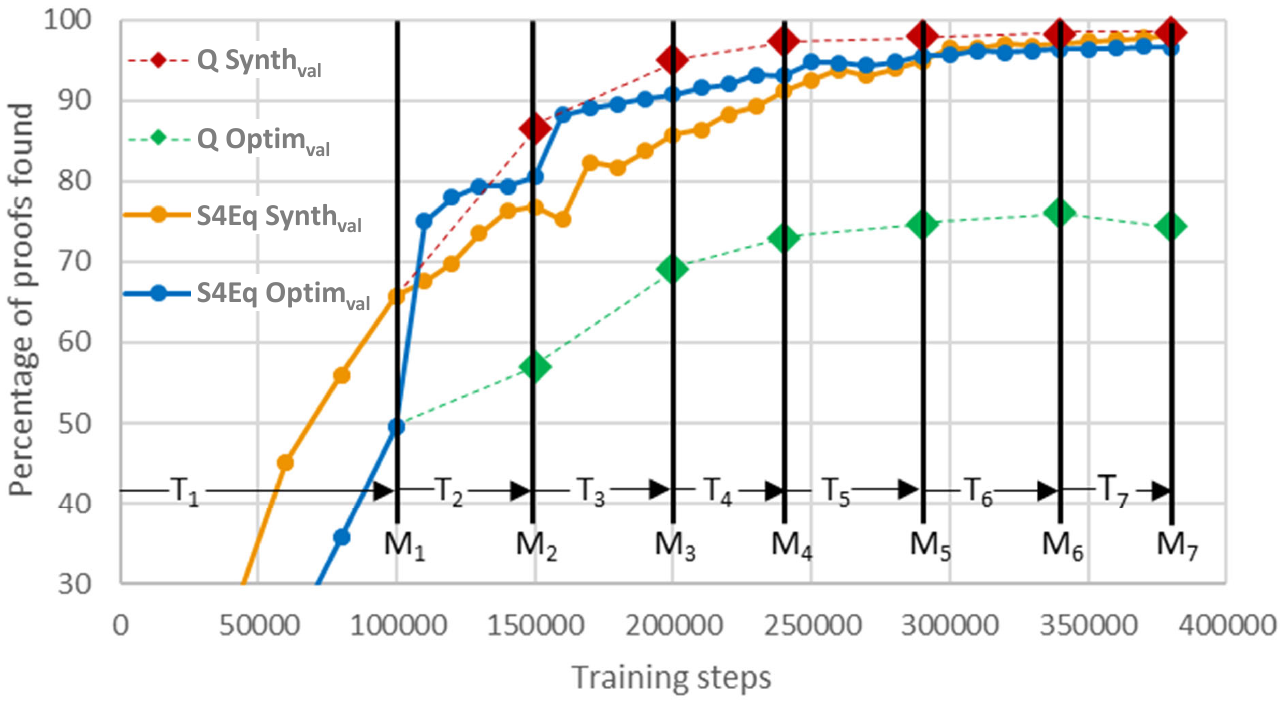}
\caption{Performance metrics through training process. $M_7$ is the \DLTOOL final model, $Q$ continues training only with $T_1$ supervised training data.}
\vspace{-.2cm}
\label{fig:TrainTune}
\end{figure}

In order to show how the an initial model improves with self-supervised sample selection, we compare \DLTOOL training up to model $M_7$ against a model trained only on the supervised training set $T_1$. 
Let us first consider training results on the $Optim$ dataset, which demonstrates how \DLTOOL can generalize into new program distributions. In Figure~\ref{fig:TrainTune}, the green squares indicate the percentage of proofs found on the $Optim_{val}$ dataset as the model trains only in a supervised manner on $T_1$ (the initial supervised training set of synthetic programs with known rewrite rules). The blue circles represent training with self-supervised sample selection. With traditional training, we see the best performance of the $Q$ model on $Optim_{val}$ occurs at 340,000 total training steps (\DLTOOL had trained to $M_6$ after that many steps). This result is significantly below the performance \DLTOOL achieves. The significant improvement of \DLTOOL on the $Optim_{val}$ dataset demonstrates the benefits on our incremental training procedure.

After training from 340,000 steps to 380,000 steps, $Q$ decreased performance on the GitHub compiled test dataset. This indicates that $Q$ was starting to overfit on the distribution for $Synth$ and the latest learnings were not as applicable to the $Optim$ problems.
On the contrary, since self-supervised sample selection generates new training samples, \DLTOOL is able to avoid overfitting and continues to improve the overall model's ability to find proofs on the $Optim_{val}$ dataset.

The $T_1$ training data is derived from the $Synth$ dataset and we see that the $Q$ model does improve performance on this dataset as it trains. After 380,000 steps, $Q$ was able to slightly outperform $M_7$ on the $Synth_{val}$ dataset (98.96\% pass versus 98.44\%) which shows that the $T_1$ dataset had sufficient samples to train for the problem distribution in $Synth_{val}$ without overfitting that dataset.

\begin{table}[h!tb]
  \caption{Performance of $M_1$, $Q$, and $M_7$ models on the 10,000 $Optim_{test}$ test pairs based on GitHub code. The self-selected samples in $T_{2-7}$ are biased to areas where $M_1$ is weak, ultimately allowing $M_7$ to outperform $Q$ in all categories. \vspace{-.1cm} }

  \label{tab:M1M1ppM7}
  \scriptsize
  \setlength\tabcolsep{4pt}
  \centering
  \begin{tabular}{@{}lrrrrrrr@{}}
    \toprule
    Sample or Proof & & $T_1$ & $T_{2-7}$ & & $M_1$ & $Q$ & $M_7$ \\
    Sequence Used & & Samples & Samples & & Proved & Proved & Proved   \\
    \cmidrule{1-1} \cmidrule{3-4} \cmidrule{6-8}
    Any rewrite rule & & 640,000 & 714,332 & & 4,866 & 7,531 & 9,688 \\
    Rename & & 5,267 & 54,925 & & 2,591 & 4,791 & 6,315 \\
    Newtmp & & 7,706 & 33,053 & & 545 & 810 & 2,277 \\
    DistributeLeft & & 30,029 & 24,243 & & 526 & 649 & 774 \\
    No statement rules & & 506,217 & 479,637 & & 968 & 1,109 & 1,111 \\
    NodeID at depth 5 & & 5,969 & 35,842 & & 18 & 91 & 323 \\
    Rewrite steps 1-10 & & 367,976 & 336,344 & & 4,690 & 7,329 & 8,997 \\
    Rewrite steps 11+ & & 272,024 & 377,988 & & 176 & 202 & 691 \\
    \bottomrule
  \end{tabular}
\end{table}

Table~\ref{tab:M1M1ppM7} shows the benefit of training with samples that are sampled from challenging proofs. The first row of the table summarizes the total number of samples available in $T_1$ used to train $M_1$ and $Q$, the total number of samples in $T_{2-7}$ used to train incrementally up to $M_7$, and the total proofs in the 10,000 sample GitHub test dataset $Optim_{test}$ found by the models $M_1$, $Q$ and $M_7$. For example, in the upper right corner we show that there were 9,688 (out of 10,000) GitHub program pairs for which proofs were found by the best model $M_7$.
Different subsets of these 9,688 proofs are shown in later rows regarding the rewrite rules needed to prove the GitHub program compilation cases.
The 2nd-4th rows introduce the mechanism through which self-supervised sample selection most benefits the model. These rows summarize the number of proofs in which at least one step used the given rewrite rule and, hence, the rows can give an indication about how well each model has learned to apply the given rule in multistep proofs. For the \texttt{Rename} rewrite rule, we see that, of the 9,688 samples proven equal by $M_7$, 6,315 of them used \texttt{Rename} for at least one step, but $M_1$ only used \texttt{Rename} in 2,591 of its proofs (less than half of the $M_7$ usage). We can see that from the $T_1$ and $T_{2-7}$ columns that self-supervised selection recognized that $M_1$ was weak in this area (implying searches with $i_{easy}$ rarely found the proof) but was able to augment the training data with more samples using the \texttt{Rename} rule (implying searches with $i_{hard}$ were able to provide example successes). We can see that model $Q$, which continued training only with $T_1$ samples, did not learn this rule as well as $M_7$. We see a similar effect for the \texttt{Newtmp} rule; 2,277 proofs used this rule for $M_7$, but only 545 used it for $M_1$. We see that $T_{2-7}$ included more of these samples to help improve the model. As a counterexample, we see that \texttt{DistributeLeft} is used in 774 successful proofs of the high-performing $M_7$ model, and in 526 of the proofs for the initial $M_1$ model. Given that $M_7$ was able to solve 97\% of the problems in $Optim_{test}$, $M_1$ had already solved over two thirds of these cases which implies $T_{2-7}$ only needed to contain certain cases of \texttt{DistributeLeft} where the model was still challenged in order to improve the model. In this way, self-supervised sample selection provides new training samples to improve the areas in which the model is weak.

Table~\ref{tab:M1M1ppM7}  also includes 'NodeID at depth 5', which indicates a rewrite rule was used on an AST node at depth 5 of an expression. Since we use NodeIDs to identify rule application positions, there are $2^{5-1} = 16$ different IDs needed to correctly locate a given instance. Fewer than 1\% of $T_1$ samples (5969) include a depth 5 node, which correlates with poor base model performance (18 cases proven). However, part of our incremental training data are the hindsight steps which use a depth 5 node. Consequently, the performance increases to 323 found proofs, meaning that the self-supervised training greatly improves the results.

\begin{figure*}
\vspace{-.1cm}
\centering\includegraphics[width=18.25cm]{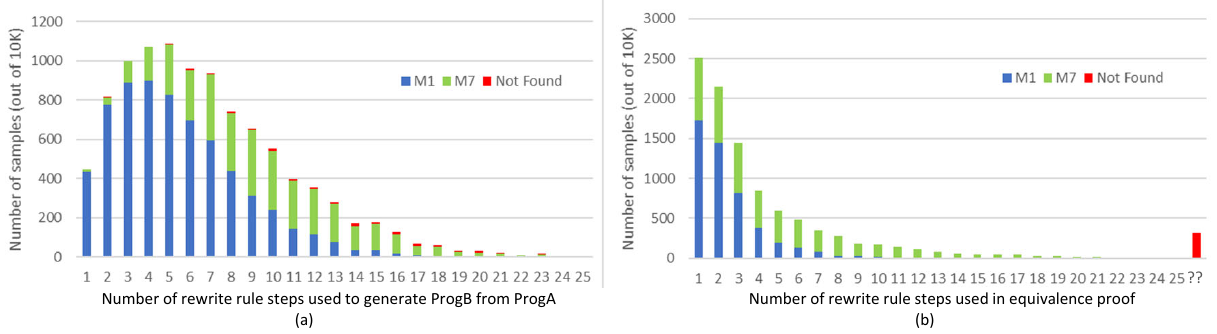}
\caption{Proofs found on test sets $Synth_{test}$ and $Optim_{test}$. Rewrite steps needed by $Optim$ samples are not known at test time and must be discovered by the model. \revision{For the $Synth$ samples, the known steps to create a program pair allow failures to be shown in the proof length column; for the $Optim$ samples, M1 vs M7 successes are shown in the column for the M7 proof length - the ?? column denotes the 312 samples for which M7 could not find a proof hence the length is not known.}}

\vspace{-.4cm}
\label{fig:Proofs}
\end{figure*}

Self-supervised sample selection also improves performance on long proofs by increasing the number of samples from such proofs. The last 2 rows of Table~\ref{tab:M1M1ppM7} report on the number of rewrite rule steps required by the 3 different models to prove pairs equivalent. The last column shows us that $M_7$ successfully used long proofs for 691 samples, while $M_1$ only found 176 long proofs. Indeed, we also see that $T_{2-7}$ contained more samples from long proofs than $T_1$ to help the model improve this category, explaining this difference.

In order to provide more insight into how \DLTOOL handles and reacts to challenging proofs, we now discuss the first proofs searched for with $M_1$ to create $T_2$ which was used to train $M_2$ and also the last proofs searched for by $M_6$ to create $T_7$ used to train $M_7$. Recall that our $Optim_{val}$ model evaluations are done with $i=10$, and the proof searches done for self-supervised sample selection bracket this value (the searches use $i_{easy} = 2$ and $i_{hard} = 20$). $M_1$ was able to solve 49.5\% of the $Optim_{val}$ samples; from the unsupervised program pairs $M_1$ attempted for $T_2$ generation, $M_1$ solved 45.0\% of them with $i_{easy}$ and 60.6\% with $i_{hard}$. Additionally, for 8.7\% of the pairs proven with $i_{easy}$, the proof found with $i_{hard}$ was at least 2 steps shorter. After training with the 126,630 single-step samples generated for $T_2$, $M_2$ was able to solve 80.5\% of the $Optim_{val}$ samples. Similarly, $M_6$ was able to solve 96.3\% of the $Optim_{val}$ samples; it solved 94.2\% of the unsupervised samples with $i_{easy}$ and 97.4\% with $i_{hard}$. After training with 83,045 samples in $T_7$, $M_7$ was able to solve 96.6\% of the $Optim_{val}$ samples. Note that while the absolute difference between $i_{easy}$ and $i_{hard}$ was over 15\% for $M_1$ but just over 3\% for $M_6$, the total number of single-step samples in $T_1$ was less than double the count in $T_7$; this is an effect of the challenging proofs increasing in step count as the models train. While fewer proofs using $i_{easy}$ fail with the later model, the lengths of the proofs which do fail tend to be longer.

Figure~\ref{fig:Proofs} diagrams how $M_7$ solves longer proofs better than $M_1$. Recall that $M_1$ can only prove 176 samples from $Optim_{test}$ equivalent with a proof over 10 steps. Of the 691 samples that $M_7$ solves with over 10 steps, $M_1$ only solves 23 of them. The other 153 samples for which $M_1$ found a long proof are still proven by $M_7$, but in 10 steps or fewer. Figure~\ref{fig:Proofs} shows the benefit of self-supervised sample selection on the performance of proofs of increasing length in the test datasets. Here we see that proofs for $Optim_{test}$ samples which $M_7$ proved with only 1-3 steps were solved with over 50\% success by $M_1$; yet proofs that $M_7$ found with over 10 steps were rarely solved by $M_1$.

We now assess the ability of self-supervised sample selection to avoid catastrophic forgetting by analyzing the samples which $M_1$, $Q$, and $M_7$ are able to prove. Of the 4,866 samples that $M_1$ is able to prove equivalent, \emph{ALL} of them are included in the 9,688 samples that $M_7$ proves after training based on self-supervised sample selection. However, only 4,774 of them are included in the 7,531 which $Q$ proves - $Q$ 'forgot' how to prove 92 samples that the $M_1$ model it trained from had proven. \emph{This shows clearly the benefit of self-supervised sample selection for avoiding catastrophic forgetting.}

\begin{mdframed}
Answer to RQ2: Compared to supervised training, using self-supervised sample selection improved performance on our target dataset from 75\% success (supervised training) to 97\% success (\DLTOOL's novel self-supervised training). Self-supervised sample selection does focus on areas where the model needed the most improvement by selecting the most interesting new training samples.
\end{mdframed}

\subsubsection{RQ3: Generalization Ability}
\label{sec:generalize}

\begin{figure*}
\vspace{-.1cm}
\centering\includegraphics[width=15cm]{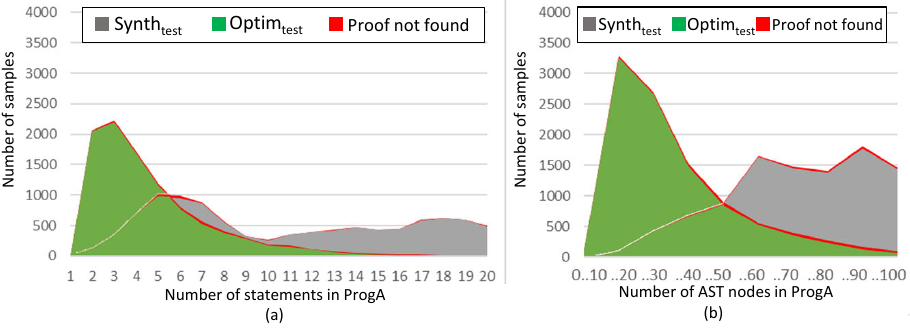}
\caption{Proofs found on test sets $Synth_{test}$ and $Optim_{test}$ plotted for varying statement and AST node counts. In all cases the performance of \DLTOOL closely tracks the dataset distribution.}

\vspace{-.1cm}
\label{fig:Passing}
\end{figure*}

A concern with machine learning is that it may learn to perform well within the samples on which it is trained but not generalize well to unseen problems that humans would consider related. We asses this risk by presenting data on how well \DLTOOL has generalized beyond its training distribution.

Let us first discuss the differences in the distribution of programs and rewrite rules between $Synth$ and $Optim$.
Figure~\ref{fig:Passing} shows the histograms of the number of assign statements and tokens in ProgA for $Synth$ and $Optim$. We see that the results of our straight-line code mining and processing of human-written code from GitHub (shown as the green distribution) tends to have many samples with fewer than 5 statements and fewer than 50 tokens, while the synthetic code (shown as the grey distribution) was designed to create more complex examples and hence has ProgAs with more statements and more AST nodes.
This clearly shows a different distribution.
The red areas show the proofs which are not found.
The thinness of the red area showing that there is no obvious weakness on any area of the ProgA distribution. The rewrite rules needed to prove equivalence also varies between $Synth$ and $Optim$: 49\% of the proofs for samples in $Synth_{test}$ are solved without rewrite rules related to statements while less than 12\% of proofs on the $Optim_{test}$ dataset could be solved without these rules (compare the ``No statement rules'' rows from tables~\ref{tab:categories} and~\ref{tab:M1M1ppM7}).
\revision{In relation to rewrite rules, there were 1,271,787 total training samples for $T_{1-6}$ and 18,216 unique rewrite rule targets; $T_7$ added 83,145 new samples including 556 new targets which were not in the previous rewrite rule group, hence indicating the ability of the model to generate previously unseen rule selections.} Taken together, these data show that \DLTOOL has generalized well over 2 different datasets.



In order to show \DLTOOL can generalize outside its training domain, we used algorithm~\ref{alg:GenerateRewrites} with CheckLimits and GenerateProgA adjusted such that we created test samples with 101-120 AST nodes and 3 outputs, which is outside the initial training distribution. Recall that on $Synth$ (with up to 100 AST nodes and 1 or 2 outputs), \DLTOOL achieves 98\% success on the $Synth_{test}$ test set.
Now, on programs with 101-120 AST nodes and 3 outputs it is still able to prove 60\% of the program pairs equivalent, showing that \DLTOOL can generalize to larger examples.

%

\begin{figure}
\vspace{-.1cm}
\centering\includegraphics[width=9cm]{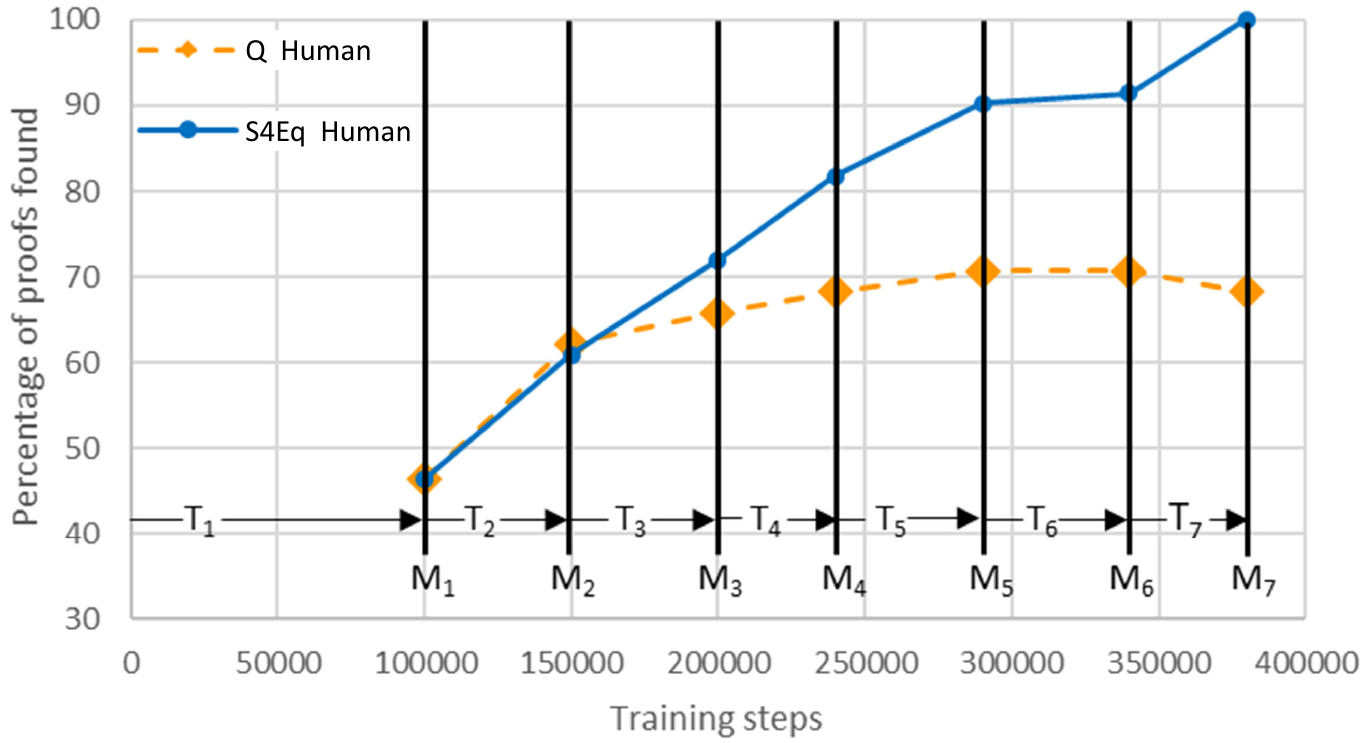}
\caption{$M_7$ searched 305,748 program pairs based on human-written GitHub functions and found 82 equivalent pairs. This chart shows the percentage of these 82 cases found as \DLTOOL trains and as Q trains (Q trains only on $T_1$ data).}
\vspace{-.2cm}
\label{fig:GHEq}
\end{figure}

\begin{table*}
\vspace{-.1cm}
  \small
  \centering
  \begin{tabular}{@{}lrrr@{}}
    \toprule
    & & \multicolumn{2}{c}{Percent proofs found} \\
    Model description & & $Synth_{val}$ & $Optim_{val}$   \\
    \cmidrule{1-1} \cmidrule{3-4}
    $M_7$ (Golden model) & & 98\% & 97\% \\
    $M_1$ (Golden model before incremental training) & & 66\% & 50\% \\
    Faster learning rate (0.0002 vs 0.0001 in $M_1$) & & 2\% & 1\% \\
    Slower learning rate (0.00005 vs 0.0001 in $M_1$) & & 49\% & 35\% \\
    Fewer transformer layers (6 vs 8 in $M_1$) & & 56\% & 23\% \\
    Limit language to scalars (vs scalars+vectors in $M_1$) & & 63\%* & 40\% \\
    Linear algebra expressions only (and 50 AST node limit vs 100 in $M_1$) & & 99\%* & 9\% \\
    \bottomrule
  \end{tabular}
  \caption{\DLTOOL ablation study. With the exception of the 'Golden model', results are shown with only initial base training of the model. The synthetic datasets for the last 2 rows were regenerated based on the modified language grammar. }\label{tab:ablation}
 \end{table*}

For our final evidence of generalization, we present equivalences found between human-written programs on GitHub.
Of the 305,748 program pairs in $Human$, we found 82 provable cases of equivalence. Figure~\ref{fig:EquivExample} shows an example of proven equivalence of GitHub samples. Figure~\ref{fig:GHEq} illustrates the generalization of \DLTOOL to this problem as training progresses by showing the percentage of the 82 proofs found by $M_1$ through $M_7$. The figure also shows the performance of $Q$ (which only trains on $T_1$ sampled from $Synth$) on the same 82 proofs. We see that after 150,000 training steps $M_2$ and $Q$ perform about equally well but ultimately $Q$ plateaus with just over 70\% of the proofs found at 340,000 steps. Like Figure~\ref{fig:TrainTune} on $Optim_{test}$ data, $Q$ seems to start overfitting on the $T_1$ data after 340,000 steps and does not generalize well to different problem areas. However, we see that \DLTOOL continued to improve on the $Human$ test set from $M_6$ to $M_7$ showing it has generalized to the problem of finding equivalence between human-written programs well.

\begin{mdframed}
Answer to RQ3: \DLTOOL is able to generalize to problem domains outside its initial supervised training dataset.
\DLTOOL is able to solve 60\% of synthetic programs that are outside of any training data it was presented with. Furthermore, it is able to progressively generalize to find up to 82 proofs of equivalence in the golden human-written samples from GitHub.
\end{mdframed}

\subsection{Ablation Study}
\label{sec:ablation}

Table~\ref{tab:ablation} illustrates some of the model variations we explored. Except for the golden model, all of the results are reported only after initial training using the appropriate synthetic dataset. All of the models are also evaluated on the GitHub dataset.

The faster learning rate shown in Table~\ref{tab:ablation} has poor results presumably due to known risks of divergence with large learning rates. We studied this result further by attempting to do an iteration with self-supervised sample selection using the poor proof success. As discussed in Section~\ref{sec:exps4}, self-supervised sample selection requires that sufficient examples of the input and output tokens are provided for successful incremental training. For example, using the $M_1$ that led to our golden model, the $T_2$ dataset had 126,630 samples for use during training, 11,737 of which used the Rename rule. For the high learning rate model, since even the $i$=20 search had poor success, its $T_2$ only had 1,482 samples in it and NONE included a successful use of Rename. Hence, unlike our golden model, incremental training on the high learning rate model did not significantly improve performance on $Synth_{val}$ nor $Optim_{val}$. We see also that a slower learning rate produced a slightly worse result than our $M_1$ result, hence we selected $M_1$ for our \DLTOOL training base.

We tested our transformer model with different numbers of attention layers, as well as different numbers of attention heads and hidden feature sizes. We show a typical result of these searches with only 6 attention layers. This parameter had a small loss for the $Synth_{val}$ success, but did not generalize as well to the $Optim_{val}$ dataset and hence it was not pursued further.

The last 2 rows of Table~\ref{tab:ablation} explore training models on alternate language grammars. For 2 these cases only the synthetic proofs found column indicates the number of proofs found for the synthetic dataset which aligns with the model description. Perhaps surprisingly we see that a language with only scalar variables and operators performs worse than $M_1$ (which has both scalars and vectors) on both its own synthetic validation set as well as on $Optim_{val}$. One possible explanation for the weakness on both the synthetic validation set with only scalars and the compiled GitHub programs in $Optim_{val}$, which only use scalars, is that the existence of vectors in the training set helps the attention layers in the model generalize better to complex uses of the rewrite rules. The final row of the ablation study shows results for a language which has only a single statement with up to 50 AST nodes, but matrixes, vectors, and scalars are supported. We see strong success on a synthetic dataset with the same features, but this model does poorly on the GitHub compiled equivalence checks. When compared with prior work on a similar dataset \cite{Kommrusch20ggnn} this row demonstrates that \emph{for our problem of program equivalence the transformer model outperforms a well tuned graph-to-sequence model which itself was found to outperform a bidirectional RNN sequence-to-sequence model}.

\begin{table}
  \caption{Results for 10,000 pairs in $Optim_{test}$ as proof search parameters are varied. Parameters for $M_7$ are shown in \textbf{bold}. \vspace{-.1cm} }

  \label{tab:searchparams}
  \small
  \setlength\tabcolsep{3pt}
  \centering
  \begin{tabular}{@{}ccclrrcllccclrr@{}}
    \toprule
    &     &     & & Proofs & Time & & & & &     &     & & Proofs & Time \\
    $l$ & $i$ & $b$ & & Found & (sec) & & & & $l$ & $i$ & $b$ & & Found & (sec) \\
    \cmidrule{1-3} \cmidrule{5-6} \cmidrule{10-12} \cmidrule{14-15}
    50 & 1  & 1 & & 87.7\% & 831 & & & & 50 & 10 & 2 & & 96.1\% & 4177 \\
    \textbf{50} & \textbf{10} & \textbf{5} & & \textbf{96.9\%} & \textbf{8964} & & & & 50 & 10 & 10 & & 96.9\% & 13723 \\
    50 & 2 & 5 & & 95.1\% & 2428 & & & & 50 & 20 & 5 & & 97.2\% & 15823 \\
    40 & 10 & 5 & & 96.8\% & 8288 & & & & 60 & 10 & 5 & & 96.9\% & 9720 \\
    \bottomrule
  \end{tabular}
\end{table}

In addition to the hyperparameters related to model selection and model training, \DLTOOL usage relies on parameters related to equivalence proof search. The three main parameters are $b$ (neural beam width), $i$ (intermediate programs), and $l$ (search step limit), they all affect both search success and search time as shown in Table~\ref{tab:searchparams}. The entry with $i=1$ and $b=1$ gives a sense for the quality of our final model: taking a single proposed rewrite rule from the neural network with only a single intermediate program still finds 87.7\% of the proofs for the pairs from $Optim_{test}$ when searching for $l=50$ steps. This result shows that a significant majority of rules produced by the model are useful for the equivalence proof. Also of note, we show the proof success rate for $M_7$ when $l=50$, $b=5$, but $i$ is set to 2, 10, and 20. This shows the performance and time results for $i_{easy}=2$ and $i_{hard}=20$ as well as our search $i$ setting of $i=10$. In particular, if we continued with self-supervised sample selection, 2.1\% of proofs are found with $i_{hard}$ but not $i_{easy}$, allowing for continued proofs on which to improve the model. The other entries in the table show that we selected our proof search parameters near the point where further time did not significantly improve the percentage of proofs found.

This ablation study does not include the full breadth of models and language representations we explored. For our transformer interface, we tested input variations (such as using infix instead of prefix or including statement numbers in the input) and output variations (such as different rewrite rule syntax including left/right path listing to identify expression nodes and also outputting the full rewrite sequence as a single long output). We also tested sequence-to-sequence RNN and graph-to-sequence models on early versions of our language \cite{Kommrusch20ggnn}, and we explored transformer model parameters guided by OpenAI's work on neural language model scaling \cite{Kaplan20Scaling}. This testing, along with our results showing improved success from $M_1$ through to $M_7$ demonstrate that a machine learning model can learn to find rewrite rules proving equivalence effectively, and in the end the parameters for our golden transformer model performed best overall in these studies.

\subsection{Execution time}

Table~\ref{tab:time} shows the cumulative machine hours needed for the key steps related to $M_7$ training and usage. 

\begin{table}[h!tb]
  \caption{Execution time statistics on \DLTOOL tasks. Machine hours are approximate as the systems are pre-emptable by students.\vspace{-.1cm} }

  \label{tab:time}
  \footnotesize
  \setlength\tabcolsep{3pt}
  \centering
  \begin{tabular}{@{}lrr@{}}
    \toprule
    & & Approx \\
    & & Machine \\
    Task Description & & Hours \\
    \cmidrule{1-1} \cmidrule{3-3}
    Train 16 $M_1$ candidates with varying hyperparameters & & 375 \\
    Search for easy and hard proofs on total of 600,000 & & 540 \\
    \hspace{0.2cm} equivalent pairs with models $M_1$ through $M_6$ & & \\
    Train 4 candidates each for models $M_2$ through $M_7$ & & 270 \\
    \hspace{0.2cm} with varying learning rates and learning rate decays & & \\
    \cmidrule{1-1} \cmidrule{3-3}
    Total to create $M_7$ with self-supervised sample selection & & 1185 \\
    Total to create $Q$ with only supervised samples & & 645 \\
    \cmidrule{1-1} \cmidrule{3-3}
    Search 305,748 mostly unequal abstractions of & & 775 \\
    \hspace{0.2cm} human-written code on $M_7$ with $b=5, i=10, l=50$ & & \\
    \bottomrule
  \end{tabular}
\end{table}

For some steps we use up to 30 machines in parallel as indicated in Section~\ref{sec:expprotocol}. This table shows that while our proof search for self-supervised sample selection takes time, it does not double the model creation time for $M_7$ relative to a model trained with traditional hyperparameter searches such as $Q$. Also of note is that almost all of the $Human$ pairs are not equal, and the average search time per pair with a 50-step limit takes about 9 seconds. Meanwhile, as seen in Table~\ref{tab:searchparams}, the equivalent pairs in $Optim_{test}$ only take about 0.9 seconds per pair to search on average because once the proof is found the search terminates.

\subsection{\revision{Comparison to exhaustive search baseline}}

\label{subsec:comparisontoexhaustive}

\revision{Table~\ref{tab:baseline} presents proof search data, for samples from $Synth_{test}$ of varying sizes and varying rewrite rule lengths between ProgA and ProgB. An exhaustive search, by breadth-first search traversal of the Program Equivalence solution Space graph ($PES$), is compared against \DLTOOL. Table~\ref{tab:baseline} illustrates the in-practice complexity of this equivalence problem, in relationship to program size and proof length. Note \DLTOOL handles proofs of 50 steps in seconds. }

\revision{Search paths explored in the $PES$ of progA are reported, and the number of unique programs constructed during the search (i.e., the distinct vertices in the $PES$ visited). Expectedly there is an exponential number of vertices visited when using exhaustive BFS. Consequently, as our goal is only to illustrate this in-practice exponential (space) complexity, we report data averaged over program pairs we selected in $Synth_{test}$ to keep the execution time of the exhaustive search manageable. That is, we report averages over all program pairs in  $Synth_{test}$ for small programs with 1 and 2 steps (113 and 124 total), a random sample of only 20 programs (out of 103) for 3 steps, and 10 out of 73 for 4 steps. For large programs, we randomly sample 20 pairs (out of 690) in  $Synth_{test}$  starting 2 steps, and only 4 (out of 897) for 3 steps. The exponential in-practice space complexity is clearly visible in Table~\ref{tab:baseline}, and such exhaustive search approaches cannot scale beyond a few proof steps. }


\begin{table}[h!tb]
  \caption{\revision{Search space for $Synth_{test}$ program pairs. Exhaustive bread-first search paths and unique programs explored to find proofs with a given number of steps in generation of ProgB from ProgA. Short programs have up to 30 AST nodes while large programs have over 30.} \vspace{-.1cm} }

  \label{tab:baseline}
  \footnotesize
  \setlength\tabcolsep{4pt}
  \centering
  \begin{tabular}{@{}lrrrrrrr@{}}
    \toprule
    & & \multicolumn{6}{c}{Generation rewrite rule steps} \\
    Description & & 1 & 2 & 3 & 4 & 5 & 6   \\
    \cmidrule{1-1} \cmidrule{3-8}
    Small programs & & & & & & &   \\ 
    \hspace{0.1cm} Exhaustive search & & & & & & &   \\ 
    \hspace{0.2cm} paths explored & & 63 & 4167 & 160197 & 5132748 & N/A & N/A   \\ 
    \hspace{0.2cm} unique programs & & 63 & 2514 & 78806 & 2177643 & N/A & N/A   \\ 
    \hspace{0.1cm} \DLTOOL search & & & & & & &   \\ 
    \hspace{0.2cm} paths explored & & 1 & 6 & 24 & 50 & 88 & 104   \\ 
    \hspace{0.2cm} unique programs & & 1 & 4 & 13 & 27 & 48 & 60   \\ 
    Large programs & & & & & & &   \\ 
    \hspace{0.1cm} Exhaustive search & & & & & & &   \\ 
    \hspace{0.2cm} paths explored & & 162 & 34609 & 4895435 & N/A & N/A & N/A   \\ 
    \hspace{0.2cm} unique programs & & 162 & 18932 & 1957605 & N/A & N/A & N/A   \\ 
    \hspace{0.1cm} \DLTOOL search & & & & & & &   \\ 
    \hspace{0.2cm} paths explored & & 1 & 6  & 25 & 63 & 102 & 142   \\ 
    \hspace{0.2cm} unique programs & & 1 & 4 & 15 & 35 & 54 & 79   \\ 
    \bottomrule
  \end{tabular}
\end{table}


\revision{Having seen the intractability of exhaustive search, we see that \DLTOOL's search is efficient (rows S4eq in  Table~\ref{tab:baseline}). For program pairs that are 1 rewrite step apart \DLTOOL finds the correct rewrite rule on the first attempt from the model on average. Our search mechanism results in a somewhat linear space increase in search complexity as the distance between ProgA and ProgB increases, exclusively because we maintain a list of all distinct programs generated. Otherwise \DLTOOL as a constant space complexity: only the $n$ (10 in our experiments) different programs we maintain concurrently during the search.}

\revision{\DLTOOL was able to prove equal 99.8\% of the 4,876 large program pairs with rewrite distance up to 6, and seamlessly handles proof of lengths of 20 and above, as shown in Fig.~\ref{fig:Proofs}}.

\section{Discussion}

\subsection{Impact of Supervised Training}

We now discuss qualitative insights related to the datasets and models. We first note how well the Q model tracked $M_2$ on the problem of finding human-written equivalences in GitHub. This is a signal that further training with the supervised data may be an efficient path to a quality model. Indeed, Figure~\ref{fig:TrainTune} shows the Q model plateauing only after 340,000 iterations. We suspect that the optimal time to pause supervised training and generate self-supervised samples is before the supervised model plateaus. That would allow for model weights which have not yet been committed to the fully trained result to be available for adjustment with the self-supervised samples. However, for our problem, it may well have been more efficient to train $M_1$ for 200,000 steps instead of only 100,000.

\subsection{Threats to Validity}

We now discuss the threats to validity for our work. A first concern is the general time required to generate iterative training samples. While we did not seek to optimize this procedure, it is bound to consume compute resources which could potentially be used for further model training. If the full training distribution is available with supervised samples, then it may be that self-supervised sample selection would have minimal benefit. However, the ability of self-supervised sample selection to find proofs even shorter than the supervised proof provided may yield benefit if explored directly. Our work did not directly study the benefit of self-supervised sample selection on a single dataset; it studied the ability to extend a model into a problem domain which did not provide a ground truth for supervised learning. This leads to a second threat to validity: with moderate effort our code optimizations done to produce training samples for $Optim_{train}$ could have generated rewrite rules for supervised learning. In cases where such a process could be done, then even extending to a new problem domain is reduced to the problem of training a model with supervised samples. A third area for concern with our work would be how well it will generalize to more complex languages. Indeed, our prior work has shown that increasing program size and language complexity reduces accuracy of the model \cite{Kommrusch20ggnn}. Such explorations are of keen interest for future work.

The language we studied was a subset of the full C language syntax. A more complete equivalence proof on the C language would have to account for recursively verifying functions called (\texttt{AddMatrix} might be called by 2 code sequences from different projects, but the code would not be equivalent unless the function was too), loop transformations (loop unrolling or nested loop reordering), as well as pointer and array referencing.

We limited our search to 50 rewrite rule steps for time reasons and, hence, programs which are equivalent in our grammar and provable with our set of rewrite rules would not be found if the number of rules needed is more than 50. Additionally, other work has found that representing more complex code such as Java classes can typically require hundreds of tokens \cite{chen2019sequencer}.

Within the prefix encoding of AST language, we define a set of rewrite rules and claim that our system guarantees no false positives by design. A threat to this claim would be an incorrect implementation of our semantics-preserving rewrite rules and the validity checks done before they are applied (for example, if we had a bug in which Commute was allowed on scalar subtraction). Indeed, during early development phases, we found and fixed bugs of this type. While there is a risk here, our experience is that the risk is low now that we have reached high levels of accuracy and reviewed many proofs. 
\revision{Ultimately, however, the ability to detect and repair compilation step bugs is a key motivation of \DLTOOL. Our system learns to prove equivalent programs in part by being exposed to the distribution of programs under consideration. Compiler bugs may be arising when two programs supposed equivalent are not proved as such. }

\section{Related Work}

\subsection{Static Program Equivalence}

Algorithms for proving program equivalence restricted to specific classes of programs have been developed \cite{verdoolaege2012equivalence,alias2004recognition,barthou2002,iooss2014program}. These approaches are typically restricted to proving the equivalence of different schedules of operations, possibly via abstract interpretation \cite{schordan.14.isola,churchill2019semantic} or even dynamically \cite{bao2016polycheck}.
Popular techniques also involve symbolic execution to compare program behavior \cite{Mora18,Badihi20}. The problem of program equivalence we target may be solved by other brute-force (or heuristical) approaches, where a problem is solved by pathfinding. This includes theorem provers \cite{bertot2013interactive,isabelle-web}, which handle inference of axiomatic proofs. Program rewrite systems have been heavily investigated,  \cite{dershowitz1985computing,steffen1991data,clarke2003behavioral,visser2003model,namjoshi2000syntactic,kalvala2009program,mansky2010framework}. While semantics-preserving rewrite systems for program equivalence have been studied \cite{visser2004program,lucanu2015program,reddy1989rewriting,willsey21}, our contribution recognizes this compositional formalism is well suited to deep learning sequence generator systems. The merits of stochastic search to accelerate such systems has been demonstrated \cite{murawski2005probabilistic,herault2004approximate,gogate2012probabilistic}. The novelty of our work is to develop carefully crafted sequence generator neural networks to automatically learn an efficient pathfinding heuristic for this problem.




EqBench \cite{Badihi21} proposes a test suite of 147 pairs of equivalent C/Java programs. As they include if-conditionals, it is not immediately usable with \DLTOOL. In contrast, we mine and build tens of thousands of equivalent program pairs, using a richer set of rewrite rules for expressions and functions. Our complete dataset, including the samples extracted from GitHub, is publicly available as a program equivalence test suite \cite{KommruschS4Eq21}, providing a rich suite complementing EqBench's.

\subsection{Incremental and Transfer Learning}
\label{sec:relatedIncremental}

For incremental learning in \DLTOOL, we use an "instance incremental scenario" in that for our problem we keep the output vocabulary constant while creating new data for incremental learning model updates \cite{Yong20}. Ye \etal discuss using an output verification process (in their case compilation and test for program repair) to adjust the loss function in later training iterations \cite{Ye21}; our approach is related in that we test outputs but instead of adjusting the training loss we create new training samples which helps to generalize the model to a different problem domain. 

To the best of our knowledge, there are only a few works that use transfer learning in the software engineering domain, and none of them use it for generating equivalence proofs. Recently, Ding has done a comprehensive study on applying transfer learning on different software engineering problems \cite{Wei2021Master}, such as code documentation generation and source code summarization. He found that transfer learning improves performance on all problems, especially when the dataset is tiny and could be easily overfitted. In our work, we deploy transfer learning on program equivalence proofs and show that it also improves generalization.

Gallego \etal demonstrate the benefits of using an image recognition model in conjunction with an adversarial network to create training samples in a new domain and incrementally train image classifier neural networks  \cite{Gallego21}. Our work benefits from a specific mechanism to verify the correctness of the output generated by earlier models.

Huang, Zhou, and Chin used transfer learning to avoid the problem of having a small dataset for the error type classification task \cite{huang2020application}. They trained a Transformer model on the small dataset and achieved 7.1\% accuracy. When training first on a bigger source dataset and tuning afterward on the small dataset, they reached 69.1\% accuracy.
However, they do not develop self-supervised sample selection, and implement a limited analysis of transfer learning. In our work, we  develop a form of transfer learning for program equivalence, and carefully analyze its merits and limitations, reaching 97\% accuracy on our dataset.



\subsection{Symbolic Mathematics Using Machine Learning}
 Hussein \etal \cite{alhussein19} develop a system which learns to apply a set of inequality axioms and derived lemmas using a reinforcement learning model with a feed-forward neural network. A challenge of using reinforcement learning is the determination of a feasible reward function from the environment and the resulting training time. HOList \cite{bansal19,paliwal19} is a system that can interact with a large rule set to score tactics in the search for a proof but faces compute time limitations when training a reinforcement learning network for theorem proving. Unlike their model, we output both the tactic (rewrite rule) as well as the location to apply it (eliminating the need to score various premises individually). Similar to recent work on using transformers to propose actions \cite{chen21}, our network has learned the rewrite rules (actions on a program) which are most likely to transform the program into the target program.
Our work aims at laying the foundation for program equivalence proofs by studying a language subset that includes multiple computation statements and maintains a high accuracy.
Our approach to synthetic data modeling is similar to works by Lample \etal \cite{lample20,Lample22}, who randomly create representative symbolic equations to develop a deep learning sequence-to-sequence transformer model which can perform symbolic integration. They note that their system requires an external framework to guarantee validity. Similarly, work by Mali \etal \cite{MaliOKG21} reasons on mathematical problems but produces outputs which are not guaranteed correct. Our system outputs a verifiable reasoning that is straightforward to check for correctness, guaranteeing no false positive and the correct handling of all true negatives.

A Deep Reinforcement Learning Approach to First-Order Logic Theorem Proving \cite{Crouse21} provides the theorem prover the allowed axioms as input to the model. In contrast, we produce the rewrite rules as a sequence. As it is a reinforcement learning model, they create training iterations as proof rewards improve with better models, while our approach creates incremental samples specifically chosen through beam search to improve output by providing a correct output for a proof the model currently is challenged by.

\revision{Recently, large language models have shown promise in a variety of areas, including theorem proving as described by Polu and Sutskever \cite{Polu20}. Our work contrasts with theirs in that they pre-trained their model on 260B tokens and tuned with 32B tokens; our technique trains with less than 1B tokens. The smallest model they evaluated uses 160M parameters; \DLTOOL uses 40M parameters. They report 1,000 GPU-hours on V100's to evaluate 1,000 theorems; we report 775 GPU-hours on GTX1060's (about 1/3 the speed of V100) to attempt proofs on 305,748 program pairs. Of course their problem space is different from ours; it remains as future work to evaluate using pre-trained LLMs for program equivalence. Indeed, self-supervised sample selection for tuning such a model is an area for future study.} 

\subsection{Program Analysis using Machine Learning}
 Numerous prior works have employed (deep) machine learning for program analysis \cite{allamanisacm18,alon19,tufano19,lacomisDIRE2019,raychev2015,bavishi17}.
PHOG \cite{bielik16} presents a probabilistic
grammar to predict node types in an AST.
Program repair approaches, \cite{tufano19,chen2019sequencer} are deployed to automatically repair bugs in a program, and recent efforts in this area have found transformer models to outperform alternatives such as sequence-to-sequence models \cite{Chen22} and genetic algorithms \cite{Drain21}.
Wang \etal \cite{wang18} learn to extract the rules for Tomita grammars \cite{tomita82} with recurrent neural networks. The learned network weights are processed to create a verifiable deterministic finite automata representation of the learned grammar. This work demonstrates that deterministic grammars can be learned with neural networks, which we rely on. Recent work by Rabin \etal \cite{Rabin21} shows that neural networks (in their case GGNNs) learn more general representations of semantically equivalent programs than code2vec \cite{alon19}, which creates code representations using AST paths. Bui \etal \cite{bui20} show that using semantics preserving transformation can improve machine learning on code, and continue the work with a study on using self-supervised learning to create similar embeddings for semantically equivalent code \cite{Bui21}. Allamanis \etal \cite{Allamanis21self} trains a bug repair neural network by using self-supervision. The model learns to repair artificially created bugs using rewrite rules on non-buggy code. Ni \etal \cite{Ni22} uses a technique similar to hindsight experience replay to create new samples for training a program synthesis neural network. We use beam search to identify model weaknesses and target learning to generate the transformations that prove semantic equivalence.

\section{Conclusion}

We introduced \DLTOOL, an end-to-end deep learning framework to find equivalence proofs between two complex program blocks.
\DLTOOL emits a \emph{verified} sequence of rewrites from one program to another when successful. We have designed self-supervised sample selection, an original training technique tailored to our problem domain. This approach further improved the ability of the deep learning system to find more complex proofs, with up to 97\% success on our synthetic and GitHub-constructed evaluation sets. All our datasets are made available to the community, including synthetic generation techniques for the problem of program equivalence via rewrite rules, as well as the programs built from the ASTs we mined in C functions from GitHub.





\section*{Acknowledgments}
This work was supported in part by the U.S. National Science Foundation award CCF-1750399.
This work was partially supported by the Wallenberg Artificial Intelligence, Autonomous Systems and Software Program (WASP) funded by Knut and Alice Wallenberg Foundation, and by the Swedish Foundation for Strategic Research (SSF). Some experiments were performed on resources provided by the Swedish National Infrastructure for Computing (SNIC).

\ifCLASSOPTIONcaptionsoff
  \newpage
\fi



%

\printbibliography

%

\begin{IEEEbiography}[{\includegraphics[width=1in,height=1.25in,clip,keepaspectratio]{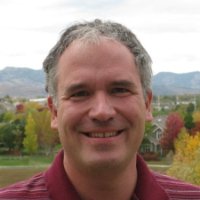}}]{Steve Kommrusch} received his PhD in computer science at Colorado State University in 2022. He received his BS in computer engineering from University of Illinois in 1987 and his MS in EECS from MIT in 1989. From 1989 through 2017, he worked in industry at Hewlett-Packard, National Semiconductor, and Advanced Micro Devices. Steve holds over 30 patents in the fields of computer graphics algorithms, silicon simulation and debug techniques, and silicon performance and power management. His research interests include Program Equivalence, Program Repair, and Constructivist AI using machine learning. More information available at: https://www.cs.colostate.edu/~steveko/.
\end{IEEEbiography}


\begin{IEEEbiography}
[{\includegraphics[width=1in,height=1.25in,clip,keepaspectratio]{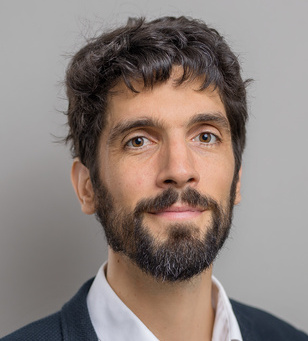}}]
{Martin Monperrus}
is Professor of Software Technology at KTH Royal Institute of Technology. He was previously associate professor at the University of Lille and adjunct researcher at Inria. He received a Ph.D. from the University of Rennes, and a Master's degree from the Compiegne University of Technology. His research lies in the field of software engineering with a current focus on automatic program repair, program hardening and chaos engineering.
\end{IEEEbiography}

\begin{IEEEbiography}[{\includegraphics[width=1in,height=1.25in,clip,keepaspectratio]{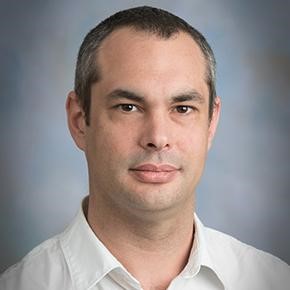}}]
{Louis-No{\"e}l Pouchet}
 is an Associate Professor of Computer Science at Colorado State University, with a joint appointment in the Electrical and Computer Engineering department. He is working on pattern-specific languages and compilers for scientific computing, and has designed numerous approaches using optimizing compilation to effectively map applications to CPUs, GPUs, FPGAs and System-on-Chips. His work spans a variety of domains including compiler optimization design especially in the polyhedral compilation framework, high-level synthesis for FPGAs and SoCs, and distributed computing. 
He is the author of the PolyOpt and PoCC polyhedral compilers, and of the PolyBench benchmarking suite.
\end{IEEEbiography}




\pagebreak

\appendix
\section{\revision{Appendix}}
\label{sec:appendix}

\revision{In Section~\ref{sec:expresults} we present the analysis of our proof search models based on 10,000 test samples in $Synth_{test}$ and 10,000 in $Optim_{test}$. We now analyze model $M_7$ (which is the result of 6 iterations of our self-supervised sample selection), model $M_1$ (the original model trained only on synthetic program pairs before self-supervised sample selection), and model $Q$ (a model created by continuing to train $M_1$ for the same number of steps as $M_7$ but with only synthetic program pairs). In this section we provide 8 specific examples of program pairs selected based on the ability of the 3 models ($M_1$, $M_7$, and $Q$) to successfully prove their equivalence.}

\revision{In Listings~\ref{lst:synM1M7} through~\ref{lst:tplnM7nQ} we present selected program pairs, their method of generation, attempted proof details from 2 models, and the proof result. The program statements use a prefix encoding. The final assignments use "===" which represents assignments to the final output variables from the programs. Variables which are not assigned before use are input variables. The generation method for synthetic programs is a randomly generated sequence of rewrite rules which transform ProgA into ProgB as per Algorithm~\ref{alg:GenerateRewrites}; the generation method for compiler equivalence programs is 1 or more compiler steps and described in Algorithm~\ref{alg:GenerateKnownEqual}. The proof attempts show either the rewrite rules used by the model to prove equivalence, or the 'most likely' sequence being investigated as the model failed (due to the beam search, up to 5 proof sequences). Finally, the proof results summarize whether the model found or failed to find an equivalence proof, the number of steps searched, the total number of axiom steps proposed by the model, the number of proposed axioms which had correct syntax, the number of proposed axioms which were legal to apply given the program structure, and finally the number of unique new intermediate programs explored during the search.}

\revision{When viewed as a group this listing collection shows the complexity of proofs and programs which our system attempts to process. Some of the cases summarized statistically in Tables~\ref{tab:categories} and~\ref{tab:M1M1ppM7} are demonstrated in these listings. For example, Table~\ref{tab:categories} shows that model M7 is challenged by programs with an AST node at depth 5 and over 30 tokens, which describes the case which M7 did not solve in Listing~\ref{lst:synQnM7}. As another example, Table~\ref{tab:M1M1ppM7} shows that M7 improves on solving long proofs relative to M1 and Q, which is illustrated by Listings~\ref{lst:synM7nQ} and~\ref{lst:tplM7nQ}. Brief comments about each listing are given in the captions.}

\pagebreak
\begin{table*}
\noindent\begin{minipage}{\linewidth}
\centering
\begin{lstlisting}[language=diff, frame=single, basicstyle=\ttfamily\footnotesize, label={lst:synM1M7}, caption={Synthetic program pair which was proven equal by M1 (and M7).}, captionpos=b, breaklines=true]
Num ProgA                                             ProgB                                            
--- --------------------                              --------------------                             
 1: s17=(*s 1s s28);                                  s17=s28;                                         
 2: v22=(nv 0v);                                      v22=(nv 0v);                                     
 3: v19=(g1v s17 v22);                                v19=(g1v s17 v22);                               
 4: v13=(h3v v19 v19);                                v13=(h3v v19 v19);                               
 5: s28=(*s s17 s17);                                 s28=(*s s17 s17);                                
 6: v16=(*v v13 s28);                                 v11=(h4v v13 v13);                               
 7: v11=(h4v v13 v13);                                v22=(*v v22 s17);                                
 8: v22=(*v v22 s17);                                 v28=(*v s17 v13);                                
 9: v28=(*v s17 v13);                                 s06=(-s s17(ns s17));                            
10: s06=(+s s17 s17);                                 v16=(*v s28 v13);                                
11: v19=(-v v19 v19);                                 v19=(-v v19 v19);                                
12: s26=(u3s s06);                                    s26=(u3s s06);                                   
13: v13=(*v 0v s28);                                  v13=0v;                                          
14: s07=(*s s26 s26);                                 s07=(*s s26 s26);                                
15: v16=v11;                                          v16=v11;                                         
16: v22=(*v v16 s07);                                 v22=(*v v16 s07);                                
17: v28=(-v v13 v13);                                 v28=(-v v13 v13);                                
18: v19=(+v v19 v19);                                 v19=(+v v19(*v 1s v19));                         
19: v22===(nv v22);                                   v22===(nv v22);                                  
20: v19===(+v v19 v28);                               v19===(+v v19 v28);                              

Num Generation              M1 attempt              M7 attempt                 Proof results            
--- ----------              --------------          --------------             -------------            
 1: stm7 Swapprev           stm1 Noop N             stm7 Swapprev              M1 proof FOUND:               
 2: stm8 Swapprev           stm6 Commute N          stm8 Swapprev               Proof steps:        9        
 3: stm9 Swapprev           stm7 Swapprev           stm9 Swapprev               Axioms Evaluated: 327        
 4: stm10 Swapprev          stm13 Multzero N        stm18 Multone Nr            Correct Syntax:   327        
 5: stm9 Flipright N        stm8 Swapprev           stm10 Swapprev              Legal Axiom:      309        
 6: stm18 Multone Nr        stm9 Swapprev           stm9 Flipright N            New ProgInt:      230        
 7: stm1 Noop N             stm10 Swapprev          stm1 Noop N                                              
 8: stm10 Commute N         stm9 Flipright N        stm10 Commute N            M7 proof FOUND:               
 9: stm13 Multzero N        stm18 Multone Nr        stm13 Multzero N            Proof steps:        9        
                                                                                Axioms Evaluated: 326        
                                                                                Correct Syntax:   321        
                                                                                Legal Axiom:      275        
                                                                                New ProgInt:      207      


                                                                    
\end{lstlisting}
\end{minipage}
\end{table*}

\begin{table*}
\noindent\begin{minipage}{\linewidth}
\centering
\begin{lstlisting}[language=diff, frame=single, basicstyle=\ttfamily\footnotesize, label={lst:tplM1M7}, caption={Compiler equivalence program pair which was proven equal by both M1 and M7.}, captionpos=b, breaklines=true]
Num ProgA                                             ProgB                                            
--- --------------------                              --------------------                             
 1: s17=(/s s17 s24);                                 s27=(/s s17 s24);                                
 2: s23=(*s s17 s23);                                 s07=(*s s27 s23);                                
 3: s17=(+s s23 s17);                                 s15=(+s s07 s27);                                
 4: s16===(/s(-s s17 s23)s04);                        s16===(/s(-s s15 s07)s04);                       

Num Generation              M1 attempt              M7 attempt                 Proof results            
--- ----------              --------------          --------------             -------------            
 1: Reuse                   stm3 Rename s15         stm1 Rename s27            M1 proof FOUND:               
 2: StrReduct               stm2 Rename s07         stm3 Rename s15             Proof steps:        3        
 3: CmnSubExp               stm1 Rename s27         stm2 Rename s07             Axioms Evaluated:  33        
 4: Template                                                                    Correct Syntax:    33        
                                                                                Legal Axiom:       19        
                                                                                New ProgInt:       19        
                                                                                                             
                                                                               M7 proof FOUND:               
                                                                                Proof steps:        3        
                                                                                Axioms Evaluated:  31        
                                                                                Correct Syntax:    31        
                                                                                Legal Axiom:       27        
                                                                                New ProgInt:       20     
\end{lstlisting}
\end{minipage}
\end{table*}

\begin{table*}
\noindent\begin{minipage}{\linewidth}
\centering
\begin{lstlisting}[language=diff, frame=single, basicstyle=\ttfamily\footnotesize, label={lst:synM7nQ}, caption={Synthetic program pair which was proven equal by M7 but NOT Q. (The 'most likely' Q proof attempt continued to the 50 step limit; the first 20 steps are shown).}, captionpos=b, breaklines=true]
Num ProgA                                             ProgB
--- --------------------                              --------------------                             
 1: s28=(f1s(is(*s 1s 1s))                            s28=(f1s(is 1s)
            (/s(+s s02 s02)(*s s02 s02)));                    (*s 1s(/s(+s s02 s02)(*s s02 s02))));
 2: v22=(+v(nv(+v(nv v20)(*v s28 v20)))               v22=(+v(nv(+v(nv v20)(*v s28 v20)))
           (nv(+v(+v v20 0v)(v1v v20))));                    (nv(-v(v1v v20)(nv(+v v20 0v)))));
 3: v20=(*v(+s 1s(v3s v22))                           v20=(+v(+v(-v 0v v22)(-v(f5v s28 s28)(f4v s28 s28)))
           (-v(+v(-v 0v v22)(f5v s28 s28))                   (-v(*v(v3s v22)(+v(-v 0v v22)(f5v s28 s28)))
              (f4v s28 s28)));                                  (*v(v3s v22)(f4v s28 s28)))); 
 4: s02=(-s(+s(*s s28 s28)1s)(+s(h5s v22 v22)s28));   s02=(+s(*s s28 s28)(-s 1s(+s(h5s v22 v22)s28)));
 5: v17===(g4v s28(h3v(*v 0v s02)                     v17===(g4v s28(h3v(*v 0s 0v)
                      (-v(*v v20 s02)v22)));                            (-v(*v v20 s02)v22)));

Num Generation              Q attempt               M7 attempt                 Proof results
--- ----------              --------------          --------------             -------------
 1: stm4 Assocright N       stm1 Noop Nll           stm1 Multone Nr            Q proof FAIL:
 2: stm5 Subzero Nrll       stm1 Multone Nr         stm3 Distleft N             Search steps:      50
 3: stm2 Commute Nrl        stm2 Flipright Nrl      stm1 Noop Nll               Axioms Evaluated:2430
 4: stm3 Distleft N         stm3 Distleft N         stm2 Commute Nrl            Correct Syntax:  2427
 5: stm3 Assocright Nlr     stm3 Noop Nl            stm2 Flipright Nrl          Legal Axiom:     2087
 6: stm3 Noop Nl            stm3 Distright Nr       stm3 Noop Nl                New ProgInt:     1342
 7: stm5 Distleft Nrl       stm3 Assocright Nl      stm5 Subzero Nrll
 8: stm1 Multone Nr         stm4 Assocright N       stm3 Distright Nr          M7 proof FOUND:
 9: stm1 Noop Nll           stm5 Subzero Nrlr       stm3 Assocright Nl          Proof steps:       14
10: stm2 Flipright Nrl      stm5 Distright Nrl      stm5 Distleft Nrl           Axioms Evaluated: 595
11: stm3 Distright Nr       stm5 Multzero Nrll      stm5 Factorleft Nrl         Correct Syntax:   595
12: stm5 Factorleft Nrl     stm2 Flipleft Nr        stm5 Cancel Nrlr            Legal Axiom:      522
13: stm5 Cancel Nrlr        stm5 Commute Nrlr       stm4 Assocright N           New ProgInt:      421
14: stm5 Commute Nrl        stm2 Flipright Nrr      stm5 Commute Nrl
15:                         stm2 Flipright Nr
16:                         stm2 Commute Nr
17:                         stm2 Flipright Nr
18:                         stm5 Commute Nrlr
19:                         stm2 Commute N
20:                         stm2 Flipright Nl
\end{lstlisting}
\end{minipage}
\end{table*}

\begin{table*}
\noindent\begin{minipage}{\linewidth}
\centering
\begin{lstlisting}[language=diff, frame=single, basicstyle=\ttfamily\footnotesize, label={lst:tplM7nQ}, caption={Compiler equivalence program pair which was proven equal by M7 but NOT Q. The 'most likely' proof attempt with Q was 23 steps long; the first 20 steps are shown. Since the attempt did not reach the full limit of 50, we know that at step 23 none of the proposed rewrite rules result in a legal new intermediate program. Also note that, since 'Template' is the last generation step, ProgB is actually the program interpreted from GitHub while ProgA is generated using common subexpression elimination, strength reduction, and variable reuse. Hence, M7 was able to revert the compiled program to the original using legal rewrite rules.}, captionpos=b, breaklines=true]
Num ProgA                                             ProgB                                            
--- --------------------                              --------------------                             
 1: s22=s29;                                          s30=s18;                                         
 2: s22=(+s 1s 1s);                                   s19=s29;                                         
 3: s22=(*s s22 s22);                                 s08=(+s 1s 1s);                                  
 4: s22=(*s s22 s22);                                 s14=(*s s30(*s s08(*s s08(*s s08 s08))));        
 5: s18=(*s s18 s22);                                 s02=(*s s19(*s s08(*s s08(*s s08 s08))));        
 6: s22=(*s s29 s22);                                 s17===(f5s s20(+s s14 s04));                     
 7: s17===(f5s s20(+s s18 s04));                      s22===(f5s s20(+s s02 s04));                     
 8: s22===(f5s s20(+s s22 s04));                                                                       

Num Generation              M7 attempt              Q attempt                  Proof results            
--- ----------              --------------          --------------             -------------            
 1: Reuse                   stm6 Rename s02         stm3 Rename s14            M7 proof FOUND:               
 2: StrReduct               stm3 Rename s14         stm2 Rename s19             Proof steps:       16        
 3: CmnSubExp               stm2 Rename s08         stm1 Rename s30             Axioms Evaluated: 681        
 4: Template                stm4 Inline s14         stm5 Rename s02             Correct Syntax:   680        
 5:                         stm4 Assocright N       stm5 Inline s22             Legal Axiom:      589        
 6:                         stm1 Rename s19         stm5 Assocleft N            New ProgInt:      548        
 7:                         stm5 Inline s22         stm3 Inline s19                                          
 8:                         stm6 Inline s22         stm4 Inline s14            Q proof FAIL:                 
 9:                         stm3 Deletestm          stm5 Inline s14             Search steps:      23        
10:                         stm3 Rename s14         stm4 Swapprev               Axioms Evaluated:1055        
11:                         stm3 Deletestm          stm3 Newtmp Nll s08         Correct Syntax:  1055        
12:                         stm3 Rename s14         stm6 Usevar s19             Legal Axiom:      571        
13:                         stm3 Newtmp Nl s30      stm6 Assocleft Nl           New ProgInt:      367        
14:                         stm3 Swapprev           stm4 Assocright N                                        
15:                         stm2 Swapprev           stm6 Commute N                                           
16:                         stm5 Usevar s19         stm6 Assocright N                                        
17:                                                 stm6 Inline s19                                          
18:                                                 stm6 Usevar s08                                          
19:                                                 stm6 Commute Nrrl                                        
20:                                                 stm6 Commute Nrr             
\end{lstlisting}
\end{minipage}
\end{table*}

\begin{table*}
\noindent\begin{minipage}{\linewidth}
\centering
\begin{lstlisting}[language=diff, frame=single, basicstyle=\ttfamily\footnotesize, label={lst:synQnM7}, caption={Synthetic program pair which was proven equal by Q but NOT M7. (The 'most likely' M7 proof attempt continued to the 50 step limit; the first 20 steps are shown). Note that, as per Table~\ref{tab:categories}, in general M7 has 92\% success on at solving long programs involving nodes at depth 5 like this one, this randomly chosen example of a failure by M7 to find a proof that Q found is a case that is known challenging for M7.}, captionpos=b, breaklines=true]
Num ProgA                                             ProgB                                         
--- --------------------                              --------------------                          
 1: s05=(*s(+s(/s s06 s29)(ns s20))s29);              s05=(*s(+s(/s s06 s29)(ns(+s 0s s20)))s29);   
 2: s06=(*s(*s(v2s 0v)(is s20))                       v15=(+v(nv 0v)(u5v s29));                     
           (/s(/s s05 1s)(/s s20 s05)));              
 3: v15=(+v(nv(*v s05 0v))(u5v s29));                 s06=(*s(*s(/s s05 1s)(*s(v2s 0v)(is s20)))
                                                             (/s s05 s20));
 4: s20=(h5s(*v(u2v s06)(*s s29 s29))v15);            s20=(h5s(*v(u2v s06)(*s s29 s29))v15);        
 5: s29=(*s(-s(+s s06 s20)(/s s20 s20))               s29=(-s(-s(*s(*s(ns s05)(+s s06 s20))s20)
           (*s(ns s05)(-s s20 s06)));                           (*s(ns s05) (+s(*s s06 s06)
                                                                               (*s s06 s20))))
                                                             (*s(*s(ns s05)1s)(-s s20 s06)));
 6: s11===(+s(-s(*s s20 s06)(/s s29 s05))             s11===(+s(-s(*s s20 s06)(/s s29 s05))
             (v3s(+v 0v v15)));                                (v3s(+v 0v v15)));

Num Generation              Q attempt               M7 attempt                 Proof results
--- ----------              --------------          --------------             -------------
 1: stm2 Assocleft N        stm1 Addzero Nlrl       stm5 Flipright Nlr         Q proof FOUND: 
 2: stm3 Swapprev           stm2 Flipright Nr       stm1 Addzero Nlrl           Proof steps:       16
 3: stm3 Flipright N        stm3 Multzero Nll       stm2 Assocright N           Axioms Evaluated: 687
 4: stm5 Assocleft N        stm5 Distleft N         stm2 Assocleft Nr           Correct Syntax:   687
 5: stm5 Commute Nl         stm5 Assocleft Nr       stm2 Commute N              Legal Axiom:      605
 6: stm5 Assocright N       stm5 Assocleft Nl       stm2 Flipright Nl           New ProgInt:      508
 7: stm5 Cancel Nrlr        stm5 Commute Nll        stm2 Assocright Nl
 8: stm1 Addzero Nlrl       stm5 Cancel Nrll        stm2 Assocright N          M7 proof FAIL:
 9: stm2 Multzero Nll       stm5 Commute Nrl        stm2 Commute N              Search steps:      50
10: stm3 Commute Nl         stm5 Distright Nl       stm2 Assocright Nl          Axioms Evaluated:2430
11: stm5 Assocleft N        stm5 Assocright Nlr     stm2 Commute Nl             Correct Syntax:  2430
12: stm5 Distright Nl       stm5 Distleft Nlrr      stm2 Assocright Nl          Legal Axiom:     2164
13: stm5 Distleft N         stm5 Commute Nlrrr      stm2 Assocright N           New ProgInt:     1630
14: stm5 Distright Nl       stm2 Assocleft N        stm2 Commute N
15: stm5 Assocright Nlr     stm3 Swapprev           stm2 Commute Nl 
16: stm5 Distleft Nlrr      stm3 Commute Nl         stm2 Assocright N
17: stm5 Commute Nlrrr                              stm2 Commute N
18:                                                 stm2 Assocright Nl
19:                                                 stm2 Assocright N 
20:                                                 stm2 Commute N
\end{lstlisting}
\end{minipage}
\end{table*}

\begin{table*}
\noindent\begin{minipage}{\linewidth}
\centering
\begin{lstlisting}[language=diff, frame=single, basicstyle=\ttfamily\footnotesize, label={lst:tplQnM7}, caption={Compiler equivalence program pair which was proven equal by Q but NOT M7. While in general M7 is much better at solving compiler equivalence pairs than Q, this is one of 2 cases out of 10,000 test pairs where Q solved such a case that M7 did not. As we see from the successful proof by Q, the M7 attempt did not discover that 'stm2 Commute Nr' was all that was needed to correctly transform statement 2 and got off track with the initial attempt at 'stm2 Commute N'.}, captionpos=b, breaklines=true]
Num ProgA                                             ProgB
--- --------------------                              --------------------                             
 1: s21=(+s(*s(u4s s05)(u5s s25))                     s04=(u4s s18);                              
           (+s(*s(u4s s10)(u5s s21))
              (*s(u4s s18)(u5s s19))));
 2: s02=(+s(*s(u4s s05)(u5s s13))                     s21=(+s(*s(u4s s05)(u5s s25)) 
           (+s(*s(u4s s10)(u5s s23))                         (+s(*s s04(u5s s19))(*s(u4s s10)
              (*s(u4s s18)(u5s s02))));                                             (u5s s21))));    
 3: s12=(+s(*s(u4s s05)(u5s s17))                     s02=(+s(+s(*s(u4s s05)(u5s s13))
           (+s(*s(u4s s10)(u5s s26))                            (/s(u4s s10)(is(u5s s23))))
              (*s(u4s s18)(u5s s12))));                      (*s s04(u5s s02))); 
 4: s29=(/s(u5s 1s)s29);                              s12=(+s(+s(*s(u4s s05)(u5s s17))
                                                                (/s(*s 1s(u4s s10))(is(u5s s26))))
                                                             (*s s04(u5s s12)));                  
 5: s14=(-s s14(*s s21 s29));                         s29=(/s(u5s 1s)s29);
 6: s22===(-s s16(*s s02 s29));                       s14=(-s s14(*s s21 s29));
 7: s27===(+s s14(-s s15(*s s12 s29)));               s22===(-s s16(*s s02 s29));
 8:                                                   s27===(+s s14(-s s15(*s s12 s29)));

Num Generation              M7 attempt              Q attempt                  Proof results
--- ----------              --------------          --------------             -------------
 1: Reuse                   stm1 Newtmp Nrll s04    stm1 Newtmp Nrrl s04       M7 proof FAIL:
 2: Axioms                  stm3 Assocleft N        stm2 Commute Nr             Search steps:      50
 3:                         stm3 Flipright Nlr      stm3 Usevar s04             Axioms Evaluated:2430
 4:                         stm2 Commute N          stm3 Assocleft N            Correct Syntax:  2430
 5:                         stm2 Assocright N       stm3 Flipright Nlr          Legal Axiom:     1882
 6:                         stm2 Commute N          stm4 Assocleft N            New ProgInt:     1350
 7:                         stm2 Assocright N       stm4 Flipright Nlr
 8:                         stm4 Assocleft N        stm4 Usevar s04            Q proof FOUND:
 9:                         stm4 Divone Nlrl        stm4 Multone Nlrl           Proof steps:        9
10:                         stm4 Flipright Nlr                                  Axioms Evaluated: 331
11:                         stm2 Commute N                                      Correct Syntax:   331
12:                         stm2 Assocright N                                   Legal Axiom:      320
13:                         stm2 Commute N                                      New ProgInt:      234
14:                         stm2 Assocright N
15:                         stm2 Commute Nl
16:                         stm2 Commute N
17:                         stm2 Assocright N
18:                         stm2 Commute N
19:                         stm2 Assocright N
20:                         stm2 Commute N
\end{lstlisting}
\end{minipage}
\end{table*}

\begin{table*}
\noindent\begin{minipage}{\linewidth}
\centering
\begin{lstlisting}[language=diff, frame=single, basicstyle=\ttfamily\footnotesize, label={lst:synnM7nQ}, caption={Synthetic program pair which neither M7 nor Q proved equal.  (The ’most likely’ proof attempts continued
to the 50 step limit; the first 20 steps are shown).}, captionpos=b, breaklines=true]
Num ProgA                                              ProgB                               
--- --------------------                               --------------------                             
 1: v29=(-v(+v v20(f1v s03 s15))                       v29=(-v(-v(f1v s03 s15)(nv v20))
           (g5v(-s s09 s09)(+v(*v s15 v20)(nv v20))));        (g5v 0s(+v(*v s15 v20)(nv v20))));
 2: s07=(-s(+s(h1s v29 v29)s27)s27);                   s07=(+s(h1s v29 v29)(-s s27 s27));
 3: v20=(*v v29 s15);                                  v20=(*v v29 s15);
 4: s03=(ns(h4s(+v(nv v20)(*v s09 v29))v29));          s03=(ns(h4s(+v(nv v20)(*v s09 v29))v29));
 5: s09=(g2s(-s(*s s07 s15)(ns s27))(*v(is s03)v29));  s09=(g2s(-s(*s s07 s15)(ns s27))(*v(is s03)v29));
 6: s15=(/s(is(-s s03 s07))s09);                       s15=(/s(is(-s s03 s07))s09);
 7: s27=(+s(-s(ns s07)(+s s09 s15))                    s27=(+s(-s(ns s07)(+s s09 s15))
           (f2s(*s s15 s15)s03));                             (f2s(*s s15 s15)s03));
 8: v20===(-v(*v(-v(*v s03 v29)(*v s27 v20))           v20===(-v(*v s03(*v s07(-v(*v s03 v29)
                (*s s03 s07))                                                    (*v s27 v20))))
             (*v(*v(*v 0v 1s)s15)s09));                         (*v(*v 0v(*s s15 1s))s09));

Num Generation              Q attempt               M7 attempt                 Proof results
--- ----------              --------------          --------------             -------------
 1: stm1 Commute Nl         stm1 Flipright Nl       stm1 Cancel Nrl            Q proof FAIL:
 2: stm8 Assocright Nrl     stm8 Distleft Nl        stm1 Commute Nl             Search steps:      50
 3: stm8 Commute Nrlr       stm1 Cancel Nrl         stm1 Flipright Nl           Axioms Evaluated:2430
 4: stm8 Commute Nl         stm8 Assocright Nll     stm2 Assocright N           Correct Syntax:  2405
 5: stm1 Flipright Nl       stm8 Assocleft Nllr     stm8 Commute Nr             Legal Axiom:     2024
 6: stm1 Cancel Nrl         stm8 Commute Nllr       stm8 Multzero Nrrl          New ProgInt:     1092
 7: stm2 Assocright N       stm8 Assocright Nlr     stm8 Multone Nr
 8: stm8 Assocright Nl      stm8 Assocleft Nllr     stm8 Multzero Nrrr         M7 proof FAIL:
 9:                         stm1 Commute Nrrl       stm8 Commute Nr             Search steps:      50
10:                         stm8 Multzero Nrll      stm8 Commute Nl             Axioms Evaluated:2425
11:                         stm1 Commute Nrrl       stm8 Assocright Nr          Correct Syntax:  2425
12:                         stm1 Commute Nrr        stm8 Commute Nrr            Legal Axiom:     1562
13:                         stm8 Assocright Nllr    stm8 Commute Nl             New ProgInt:      938
14:                         stm8 Assocleft Nlrr     stm8 Commute Nr
15:                         stm1 Commute Nrr        stm8 Commute Nl
16:                         stm8 Assocleft Nllr     stm8 Assocright Nl
17:                         stm8 Assocleft Nlr      stm8 Distright Nlr
18:                         stm8 Commute Nlr        stm8 Assocleft Nlrl
19:                         stm8 Assocleft Nlrr     stm8 Commute Nlrll
20:                         stm8 Assocleft Nlr      stm8 Assocright Nlrl
\end{lstlisting}
\end{minipage}
\end{table*}

\begin{table*}
\noindent\begin{minipage}{\linewidth}
\centering
\begin{lstlisting}[language=diff, frame=single, basicstyle=\ttfamily\footnotesize, label={lst:tplnM7nQ}, caption={Compiler equivalence program pair which neither M7 nor Q proved equal. (The ’most likely’ proof attempts continued to the 50 step limit on M7 and no unique new programs were created after step 27 with Q; the first 20 steps are shown).}, captionpos=b, breaklines=true]
Num ProgA                                             ProgB                                            
--- --------------------                              --------------------                             
 1: s10=s19;                                          s06=s13;                                         
 2: s10=(+s 1s 1s);                                   s11=s19;                                         
 3: s15=(*s(ns s13)s10);                              s10=(+s 1s 1s);                                  
 4: s26=(+s 1s s10);                                  s15=(*s(ns s06)s10);                             
 5: s30=(*s s13 s26);                                 s26=(+s 1s s10);                                 
 6: s17=s30;                                          s17=(*s s06 s26);                                
 7: s15=(+s s15 s30);                                 s20=(+s s15(*s s06 s26));                        
 8: s17=(+s s17 s30);                                 s09=(+s s17(*s s06 s26));                        
 9: s15=(+s s15 s30);                                 s05=(+s s20(*s s06 s26));                        
10: s17=(+s s17 s30);                                 s08=(+s s09(*s s06 s26));                        
11: s22=(/s s13(+s s10 s10));                         s24=(+s s10 s10);                                
12: s13=(+s s15(/s(-s s13 s22)s10));                  s03=(+s s05(-s(/s s06 s10)(/s s06(/s s24 s10))));
13: s26=(+s s17(*s s10(/s s19 s26)));                 s01=(+s s08(*s s10(/s s11 s26)));                
14: s30===(+s s13 s22);                               s30===(+s s03(/s s06 s24));                      
15: s22===(+s s26 s22);                               s22===(+s s01(/s s06 s24));                      

Num Generation              M7 attempt              Q attempt                  Proof results            
--- ----------              --------------          --------------             -------------            
 1: Reuse                   stm5 Rename s06         stm8 Rename s05            M7 proof FAIL:                
 2: StrReduct               stm6 Rename s20         stm13 Rename s01            Search steps:      50        
 3: CmnSubExp               stm6 Inline s06         stm9 Rename s08             Axioms Evaluated:2430        
 4: Template                stm7 Rename s09         stm12 Distleft Nr           Correct Syntax:  2428        
 5:                         stm3 Newtmp Nll s11     stm12 Rename s03            Legal Axiom:     1738        
 6:                         stm7 Rename s17         stm10 Rename s24            New ProgInt:     1483        
 7:                         stm3 Swapprev           stm11 Swapprev                                           
 8:                         stm2 Swapprev           stm5 Rename s20            Q proof FAIL:                 
 9:                         stm1 Rename s06         stm9 Commute N              Search steps:      27        
10:                         stm7 Rename s20         stm10 Swapprev              Axioms Evaluated:1260        
11:                         stm6 Rename s17         stm9 Swapprev               Correct Syntax:  1260        
12:                         stm9 Rename s05         stm8 Rename s09             Legal Axiom:      690        
13:                         stm9 Inline s17         stm6 Inline s20             New ProgInt:      507        
14:                         stm11 Rename s08        stm7 Rename s06                                          
15:                         stm12 Rename s24        stm7 Inline s20                                          
16:                         stm13 Rename s03        stm6 Swapprev                                            
17:                         stm12 Flipright N       stm1 Newtmp N s11                                        
18:                         stm12 Flipright Nrl     stm14 Commute N                                          
19:                         stm14 Rename s01        stm2 Rename s06                                          
20:                         stm13 Usevar s06        stm14 Assocleft Nl                
\end{lstlisting}
\end{minipage}
\end{table*}

\end{document}